\documentclass[sigconf]{acmart}
\usepackage{pifont}
\usepackage{diagbox}
\AtBeginDocument{%
  }

\setcopyright{acmlicensed}
\copyrightyear{2018}
\acmYear{2018}
\acmDOI{XXXXXXX.XXXXXXX}

\acmConference[Conference acronym 'XX]{Make sure to enter the correct
  conference title from your rights confirmation emai}{June 03--05,
  2018}{Woodstock, NY}
\acmISBN{978-1-4503-XXXX-X/18/06}
\acmSubmissionID{585}

\renewcommand\footnotetextcopyrightpermission[1]{}
\begin{document}

\title{Fusion Matrix Prompt Enhanced Self-Attention Spatial-Temporal Interactive Traffic Forecasting Framework}

\author{Mu Liu}
\affiliation{%
  \institution{Jilin University}
  \city{Changchun Shi}
  \state{Jilin Sheng}
  \country{China}}
\email{liumu23@mails.jlu.edu.cn}

\author{MingChen Sun}
\affiliation{%
  \institution{Jilin University}
  \city{Changchun Shi}
  \state{Jilin Sheng}
  \country{China}}
\email{mcsun20@mails.jlu.edu.cn}

\author{YingJi Li}
\affiliation{%
  \institution{Jilin University}
  \city{Changchun Shi}
  \state{Jilin Sheng}
  \country{China}}
\email{yingji21@mails.jlu.edu.cn}

\author{Ying Wang}
\affiliation{%
  \institution{Jilin University}
  \city{Changchun Shi}
  \state{Jilin Sheng}
  \country{China}}
\email{wangying2010@jlu.edu.cn}

\renewcommand{\shortauthors}{Trovato et al.}

\begin{abstract}
Recently, spatial-temporal forecasting technology has been rapidly developed due to the increasing demand for traffic management and travel planning. However, existing traffic forecasting models still face the following limitations. On one hand, most previous studies either focus too much on real-world geographic information, neglecting the potential traffic correlation between different regions, or overlook geographical position and only model the traffic flow relationship. On the other hand, the importance of different time slices is ignored in time modeling. Therefore, we propose a Fusion Matrix Prompt Enhanced Self-Attention Spatial-Temporal Interactive Traffic Forecasting Framework (FMPESTF), which  is composed of spatial and temporal modules for down-sampling traffic data. The network is designed to establish a traffic fusion matrix considering spatial-temporal heterogeneity as a query to reconstruct a data-driven dynamic traffic data structure, which accurately reveal the flow relationship of nodes in the traffic network. In addition, we introduce attention mechanism in time modeling, and design hierarchical spatial-temporal interactive learning to help the model adapt to various traffic scenarios. Through extensive experimental on six real-world traffic datasets, our method is significantly superior to other baseline models, demonstrating its efficiency and accuracy in dealing with traffic forecasting problems.
\end{abstract}


\maketitle

\section{Introduction}
\begin{figure}[t]
\centering
\includegraphics[width=0.48\textwidth,height=45mm,trim= 50 30 20 20]{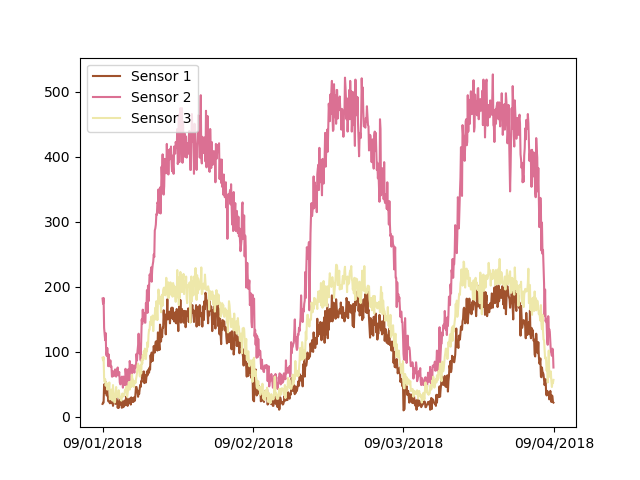}
\caption{Temporal correlations of traffic data. }
\label{periodicity}
\end{figure}

As a crucial branch of spatial-temporal prediction, traffic forecasting plays a pivotal role in numerous vital practical applications, including intelligent transportation systems and road network management \cite{saeed2023graph, zhe2020traffic}. With the rapid evolution of machine learning, deep learning techniques, particularly Convolutional Neural Networks (CNNs) \cite{claus1998evaluation} and Recurrent Neural Networks (RNNs) \cite{wojciech2014recurrent,martino2018multi-turn}, have been extensively employed in the field of traffic prediction. Recently, the emergence of Graph Neural Networks (GNNs) \cite{scarselli2009the,hao2024macro} has garnered significant attention for their application in traffic prediction tasks, resulting in remarkable achievements. Most recent research endeavors aim to harness the intricate network structure to mine abundant information from traffic data, explore complex spatial-temporal relationship patterns within it, and ultimately achieve precise and efficient forecasting. However, despite these advancements, existing methods still encounter several limitations that hinder their effectiveness in practical applications.

\textbf{Limitation1 (Spatial Dependency):} In the intricate urban road network, the traffic flow of a particular road segment is inherently influenced by the flow of other interconnected segments. This spatial dependency serves as a foundational premise for the modeling of accurate traffic forecasting. However, there are limitations in current research. Some studies narrowly focus on the influence of physical geographical locations, failing to capture the nuanced patterns of flow relationships. Conversely, others excessively prioritize potential flow impacts, neglecting the intricate dynamics of real-world road traffic conditions.

\textbf{Limitation2 (Temporal Dependency):} The traffic flow information within a designated region often exhibits analogous patterns across varying timeframes, including the morning and evening rush hours on diverse weekdays. This can be illustrated in Figure \ref{periodicity}, demonstrating the traffic flow of three observation points on three consecutive days. This temporal correlation constitutes a pivotal basis for the development of time series modeling in the realm of traffic forecasting. While current research predominantly centers on the influence of specific time periods or potential regularities, there remains a notable shortcoming in effectively integrating both long-term and short-term traffic patterns.

Therefore, we propose \textbf{F}usion \textbf{M}atrix \textbf{P}rompt \textbf{E}nhanced self-attention \textbf{S}patial-\textbf{T}emporal interactive traffic forecasting \textbf{F}ramework FMPESTF, which integrates three Spatial-Temporal Complex  modules (ST-Comp) consisting of the Fusion-Graph components and Att-Conv components to achieve more effective and generalizable traffic flow forecasting. 

To address the first limitation, we propose a Fusion-Graph module based on a fusion matrix which is formulated by the integration of a dynamic graph generation matrix and an adjacency matrix containing real traffic information. We model the dynamic traffic from two dimensions of long and short term respectively, where the long-term node pattern is stored by a special database, and the input is used as a query to calculate the spatial similarity between input representations and the node pattern, thereby facilitating the generation of the dynamic matrix. By amalgamating the dynamic and static matrices, the fusion matrix realizes the spatial information modeling of dynamic traffic and static geographic information, which enhance the capture of spatial information.

To address the second limitation, we propose a Temporal Attention module based on Attention mechanism enhanced Convolutional neural networks (Att-Conv). In the Att-Conv modules, We use attention scores for prompt learning from raw time series data to make the model pay more attention to valuable historical data for forecasting.

Our contributions are mainly summarized as follows:
\begin{itemize}
\item We propose a fusion matrix to capture intricate interaction patterns across various spatial levels in spatial-temporal data, and leverage the adjacency matrix as a prompt to enhance the modeling efficiency of spatial-temporal data.
\item We integrate an attention mechanism into the time information convolution module, so that the traffic forecasting model can pay attention to more important data information, and then effectively utilize more valuable data.
\item We conduct extensive experiments on six real-world datasets and the experimental results are significantly better than many baseline models, demonstrating the effectiveness of our model.
\end{itemize}

\section{Related Work}
\textbf{Traffic Forecasting.} Traffic Forecasting is the process of predicting traffic conditions at a specific time and location in the future \cite{renhe2023spatio-temporal, yu2023causal}, utilizing observed historical traffic data to model the complex spatial-temporal dependencies inherent in the system \cite{di2023trafformer, shengnan2022learning}. Recent research has shifted towards the utilization of more advanced techniques, specifically deep learning \cite{mengzhang2021spatial-temporal} methods and graph neural networks \cite{xiaoliang2022modeling}. These methods have exhibited excellent results in modeling the intricate spatial-temporal relationships within traffic forecasting.

\noindent\textbf{Graph Neural Networks.} GNNs aim to aggregate neighboring information of nodes via a message passing mechanism and update the feature representations of these nodes \cite{jie2020graph}. The application of GNNs in traffic forecasting predominantly centers on accurately predicting future traffic flow, congestion conditions, and travel demand \cite{weiyang2024spatiotemporal}. Its ability to capture the spatial-temporal characteristics of traffic flow data allows for more precise predictions of future traffic situations.

\noindent\textbf{Graph Prompt Learning.} Graph Prompt Learning aims to provide context and clues for the model's reasoning process by supplying additional prompt information \cite{yuhao2024graphpro,zheyuan2024can,taoran2023universal, yun2023sgl-pt}. These prompts can be additional nodes, edges, or attributes in the form of graph structures, or they can be rules or strategies that guide the model's behavior. Graph Prompt Learning first proposes the use of a task token prompt and a structural symbol prompt to bridge the gap between pre-training tasks and downstream task objectives. Additionally, Sun et al. proposed a general multi-task prompt learning method that unified the format of graphics and languages by introducing prompt graphs \cite{mingchen2022gppt}. By integrating the prompt information with the input of the graph neural network model, Graph Prompt Learning can assist the model in more accurately understanding the background and objectives of the task, and extracting key information relevant to the task.

\begin{figure}[t]
\centering
\includegraphics[width=0.46\textwidth,trim=20 80 0 0]{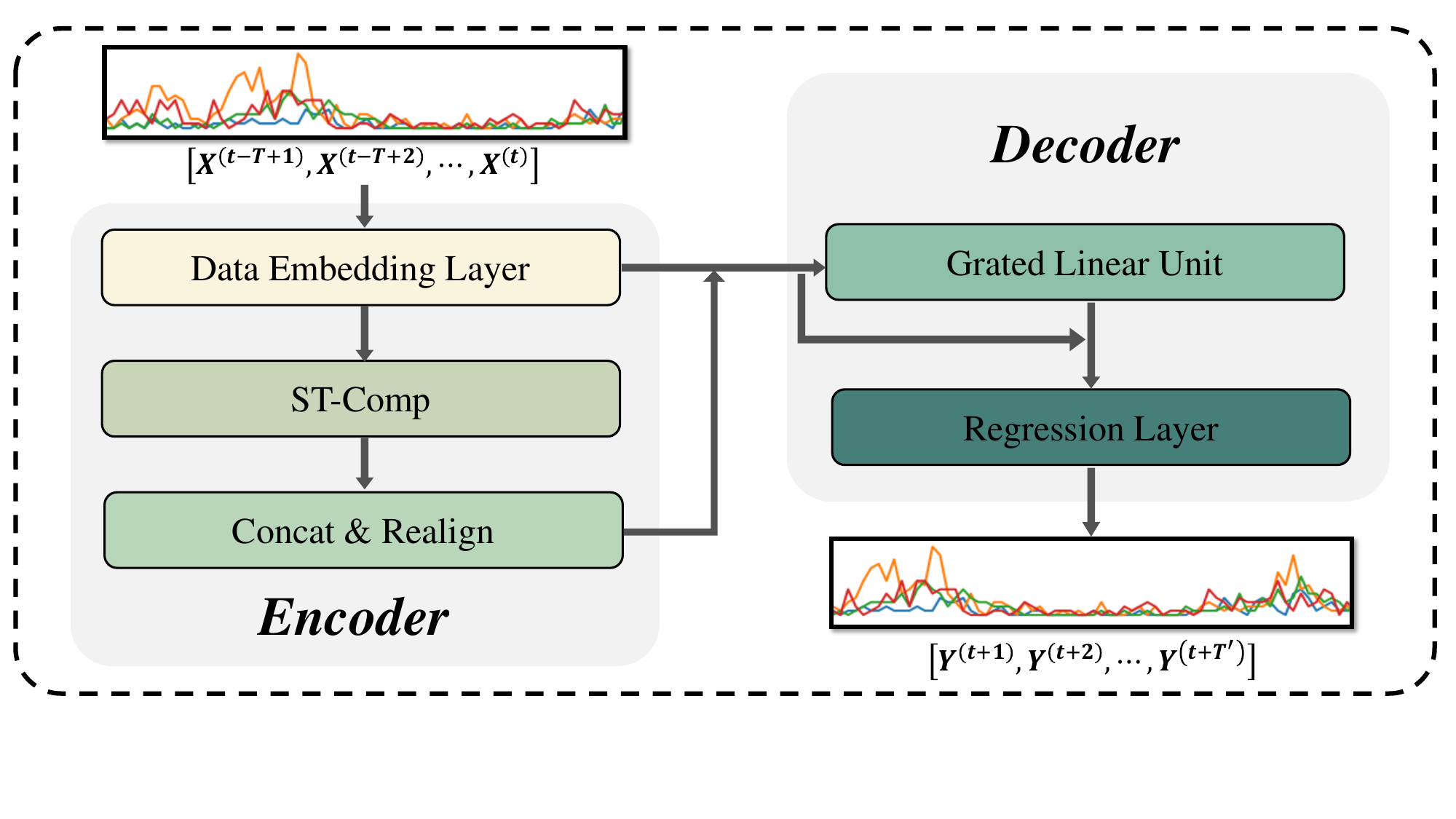}
\caption{The framework of FMPESTF.}
\label{fig1}
\end{figure}

\section{Preliminary}
In this section, we introduce the relevant definitions of traffic network information and the formalization of traffic forecasting task.

\textbf{Definition 1} (Spatial Region) According to the specific traffic pattern and road structure under the given traffic scene, we divide the scene into several geographical areas, which are defined as follows:
\begin{equation}
    V=\left \{ v_{1},v_{2},\cdots ,v_{N}  \right \} ,
\end{equation}
where \(V\) denotes spatial region set, \(N=I\times J\)denotes the region number, each element \(v_{i}\)in\(V\)denotes a spatial region.

\textbf{Definition 2} (Traffic Network) Real-world traffic networks can be described using a network of traffic sensors or through the transport links provided between stations and road segments.A traffic network can be defined as an undirected graph \(G = (V, E, A)\), where \(V\) denotes the set of nodes with the size of \(|V|=N\) , and each of node denotes an observation point in the traffic network, and \(E\) is the set of edges. \(A\in \mathbb{R} ^{N\times N} \) is the adjacency matrix of the traffic network graph \(G\), which represents the connection relationship between nodes.

\textbf{Definition 3} (Traffic Information) Traffic Information \(X^{(t)}\in \mathbb{R} ^{D\times N} \) denotes the data collected from various observation points in our planned traffic network \(G\) at time step \(t\), where \(N\) denotes the number of observation points, \(D\) denotes the number of feature channels(e.g., the demand, flow or speed).

\textbf{Problem Description} (Traffic Forecasting) The problem that traffic forecasting needs to solve is to predict the future traffic information sequence \(\left [ Y^{t+1},Y^{t+2},\cdots ,Y^{t+T^{'}}  \right ] \) using the historical sequence \(\left [ X^{t-T+1},X^{t-T+2},\cdots ,X^{t}  \right ] \), where \(T\) denotes the length of a defined sequence of historical time, \(T^{'}\) denotes the length of the time series to be predicted. The traffic forecasting task can be defined as follows:
\begin{equation}
    \left [ Y^{t+1},\cdots ,Y^{t+T^{'}}  \right ]=F\left (\left [ X^{t-T+1},\cdots ,X^{t}  \right ]  \right ) ,
\end{equation}
where \(F\) denotes the time series prediction function.

\section{Methodology}
\begin{figure*}[t]
\centering
\includegraphics[width=0.9\textwidth,trim=0 140 0 0]{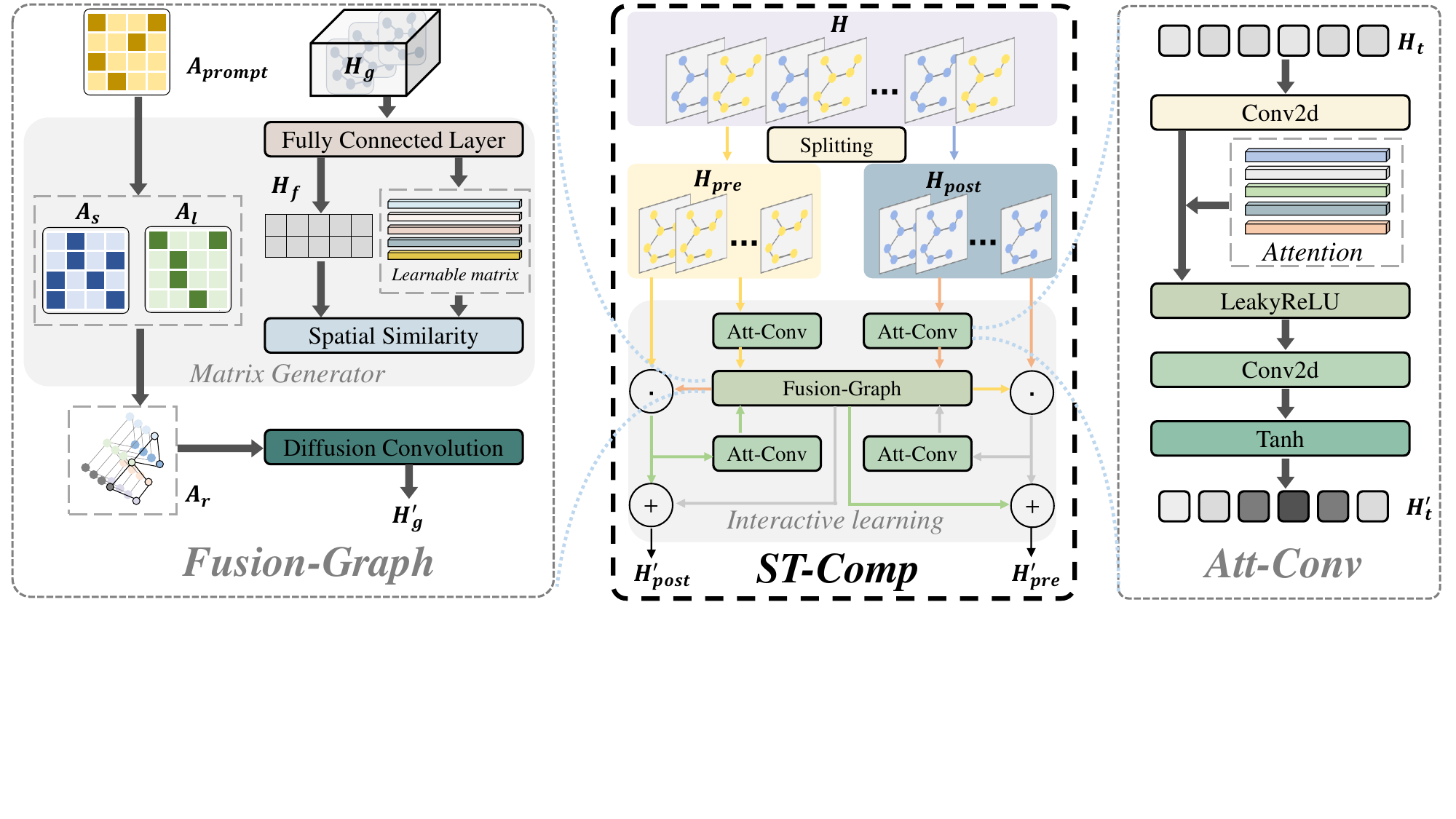}
\caption{The structure of ST-Comp module and its internal Att-Conv, Fusion-Graph components. }
\label{fig_sub}
\end{figure*}
In this section, we will introduce our fusion matrix prompt enhanced self-attention spatial-temporal interactive traffic forecasting framework and its interactive learning strategy as well as two modules: the fusion-graph module and the temporal attention module. The learning process and the overall optimization goals of our framework are as follows.

\begin{figure*}[t]
\centering
\includegraphics[width=0.9\textwidth,trim=0 320 100 20]{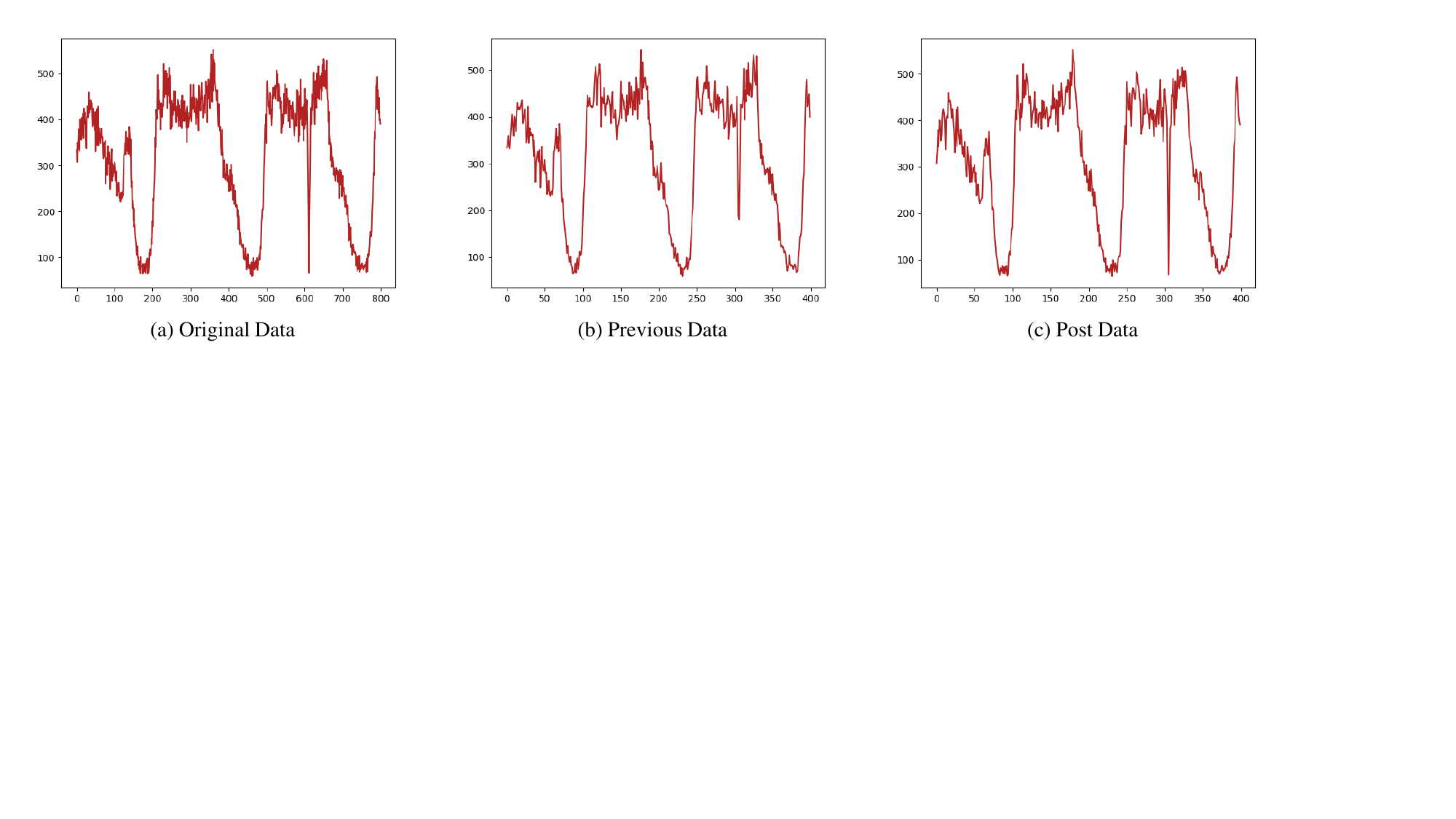}
\caption{Visualization of split traffic flow data. (a) denotes the original data, while (b) corresponds to a subsequence generated by previous exponential division, and (c) corresponds to a subsequence generated by post exponential division.  }
\label{fig2}
\end{figure*}

\subsection{Overall framework}
The framework of the proposed FMPESTF is shown in the Figure \ref{fig1}. which represents a meticulous approach to modeling complex spatial-temporal data. We leverage an intricate framework of interactive learning strategies to analyze the amassed data, subsequently fusing the distilled spatial-temporal representations for the purpose of predicting future traffic information through regression analysis. Our model embraces an encoder-decoder architecture in its entirety, ensuring a structured and comprehensive approach to the traffic forecasting task.

First, the data embedding layer, while obtaining the representation of the data collected by the sensor, integrates the periodicity and trend into it, so as to achieve the enhanced representation of the original spatial-temporal data. The high-dimensional data is then passed through multiple layers of ST-Comp modules. Each ST-Comp module contains four Att-Conv modules to learn complex temporal dependencies, and a Fusion-Graph module that introduces prompt learning technique to capture spatial dependencies and implement spatial-temporal interactive learning strategies for spatial-temporal representation interactions.

As shown in Figure \ref{fig_sub}, each ST-Comp module processes input data separately to achieve interactive learning in spatial and temporal. We use an interval sampling method to divide the input data into two subsequences of equal length. Next, the two subsequences are processed by the Att-Conv module and the Fusion-Graph module respectively, and the interaction of the two subsequences is realized in this process. In the Att-Conv module, the attention mechanism is introduced to realize the time correlation modeling, so that the model pays attention to the time slice with greater influence on the current time. In Fusion-Graph module, a learnable matrix is used to reconstruct the dynamic spatial association between nodes in the traffic network in the current period, and the adjacency matrix is introduced as the prompt learning matrix to realize the static spatial association, so as to realize the modeling of spatial information at two dimensions.

After being processed by a multi-level St module, the network generates multiple output subsequences that reintegrate these data to form a complete spatial-temporal representation. Finally, in the decoder section, the Gated Linear Unit (GLU) \cite{yann2017language} transmits these spatial-temporal representations to the regression layer to make the final prediction.

\subsection{Data Embedding Layer}
In order to excavate the deep spatial-temporal features inside the data, the original traffic data needs to be processed by the data embedding layer, and then its high-dimensional representation is obtained. First, the raw data is transformed through the fully connected layer:
\begin{equation}
    X_{raw}^{'} =Linear(X_{raw} ) ,
\end{equation}
where \(X_{raw}^{'} \in \mathbb{R} ^{d_{1} \times N\times T}\), \(X_{raw} \in \mathbb{R} ^{D\times N\times T}\), \(d_{1}\) denotes the number of expended future channels. Subsequently, in order to better learn the periodicity of data, we introduce an embedded representation with periodic information \cite{zezhi2022spatial-temporal} to incorporate prior knowledge into the model.

Traffic data exhibits periodicity and trend, such as a road section shows similar traffic conditions at the same time every week. Therefore, we embed the "day" and "week" information for each piece of data, denoted as \(X_{d}\in\mathbb{R}^{d_{2} \times N}\) and \(X_{w}\in\mathbb{R}^{d_{2} \times N}\) which contains a prior knowledge of the periodicity and trend of the data. Finally, \(H\in\mathbb{R}^{ C\times N\times T}\) is obtained by concatenating \(X_{w}\) and \(X_{d}\) with the high-dimensional representation of the raw data output of the data embedding layer:
\begin{equation}
    H=Concat(X_{raw}^{'},X_{w}+X_{d}),
\end{equation}
where \(C\) denotes the number of feature channels.

\subsection{Spatial-Temporal Component}

As a kind of time series data, traffic data is characterized by its periodicity and trend pattern in time dimension. These periodic and trend information play a key role in understanding the intrinsic time series correlation and its deep structure of the transportation system. Given this characteristic of traffic data, current traffic conditions are not only closely related to historical data, but also play an important role in predicting future traffic flows, reflecting the overall correlation of the time dimension. In the pursuit of efficient capture of temporal correlations, some approaches focus on using causal relationships between sequences for analysis, such as RNN-based and TCN-based temporal modules. However, this methodology is often limited to considering only local dependencies between current and past data, ignoring the broader global coherence of the data across the timeline. In fact, in order to capture time correlation in traffic data more comprehensively and accurately, an analytical framework that can integrate the global relationship of the series should be adopted to enhance the panoramic understanding of time dynamics, so as to achieve more accurate prediction and analysis of traffic flow changes. Inspired by previous work \cite{jingyuan2018mutilevel, minhao2022scinet, liu2024spatial-temporal}, we propose the ST- Comp module, which consists of four temporal modules that capture temporal correlation and spatial modules that capture spatial correlation, while using spatial-temporal interaction to achieve positive feedback learning between temporal modeling and spatial modeling.

As shown in Figure \ref{fig2}, we visualized the original data collected by a sensor from the PEMS08 dataset in a certain period of time and the segmented data. The subsequences retain the same periodicity and trend as the original data. As shown in Figure \ref{fig_sub}, in the ST-Comp module, we conduct spatial-temporal interactive learning of the two subsequences, in which the model maximizes the use of global relationships in the original data. In multi-level ST-Comp structures, FMPESTF can capture spatial-temporal correlations ranging from global to local perception granularity.

\noindent\textbf{Attention Convolution.} In the ST-Comp module, we propose the Att-Conv module as the temporal module to capture temporal correlation. As shown in Figure \ref{fig_sub}, each ST-Comp module uses four Att-Conv modules to capture temporal correlation. Att-Conv introduces the attention mechanism \cite{ashish2017attention} and uses two layers of 2D-CNNs, each using kernel sizes of \((1,k1)\) and \((1,k1)\), where \(k1\) and \(k2\) represent the set kernel sizes. Att-Conv processes only the time dimension of traffic data. The operation can be defined as follows:
\begin{equation}
    H^{'}_{t}=Conv2d(Attention(Conv2d(H_{t}))),
\end{equation}
where \(H_{t}\) and \(H^{'}_{t}\) denote hidden representations in Att-Conv. Through the prompt processing of attention mechanism and the two-layer convolution operation, the temporal correlation of a single sequence is effectively modeled in parallel. Combined with the periodic information incorporated in the data embedding layer, the model pays more attention to the influence of important time slices.

\noindent\textbf{Fusion Graph Generation Network.} In order to capture spatial correlation, we design a Fusion-Graph module as the spatial module. Fusion-Graph realizes two main functions of fusion graph construction and spatial association capture. In order to realize the above functions, Fusion-Graph includes a fusion matrix generator based on graph prompt and a diffusion-based GCN.

Different from the previous graph construction methods, the graph construction method adopted in fusion matrix generator not only considers the different flow patterns of each node and the dynamic correlation between nodes, but also uses the adjacency matrix as the prompt learning information. Specially, as shown in Figure \ref{fig_sub}, we define a learnable matrix \(W_{l}\in \mathbb{R}^{C\times N}\), which can adaptively store the traffic pattern of each node during training due to spatial heterogeneity. The fusion matrix generator utilizes the hidden representation \(H_{g}\in \mathbb{R}^{C\times N \times t^{'}}\) input to the Fusion-Graph, combined with the suggestion of the adjacency matrix, to construct a fusion matrix that dynamically evolves over time. This matrix not only simulates the internal relationship pattern among the nodes involved in the flow change, but also includes the real world geographical adjacency relationship.

The representation \(H_{g}\) obtained after processing by ts module is firstly nonlinear mapped by the fully connected layer in the fusion matrix generator to eliminate the hidden representation \(H_{f}\) of temporal dimension, which can be defined as follows:
\begin{equation}
    H_{f}=Linear({\textstyle \sum_{t=1}^{t^{'}}}H_{g}^{t}  ) .
\end{equation}
Next, the representation \(H_{g}\) is used as a query, and the similarity calculation is performed with the learnable matrix and itself respectively to generate two dynamic matrices, \(A_{l}\in \mathbb{R}^{N\times N }\) and \(A_{s}\in \mathbb{R}^{N\times N }\). The generation process can be defined as:


\begin{equation}
    A_{l}=SpatialSim(H_{f},W_{l}) ,
\end{equation}
\begin{equation}
    A_{l}=SpatialSim(H_{f},H_{f}) ,
\end{equation}
where \(i\) denotes the row index. The construction of \(A_{l}\) overcomes the limitation of node association in the local period, including the potential spatial association between nodes. \(A_{s}\) is the flow representation obtained by performing the similarity calculation with itself, and represents the dynamic association between nodes in the current time period. The generation of these two dynamic matrices reflects the dynamic spatial correlation among nodes from different perspectives.

Considering the potential dynamic association between nodes, static geographic information association is also an important part that cannot be ignored. Therefore, we propose to use the real world adjacency matrix \(A_{prompt}\) as a prompt to fuse the static node association information. The fusion matrix is obtained by connecting with the two dynamic matrices \(A_{l}\) and \(A_{s}\) generated above, and then fusing with the full connection layer. The generation process can be defined as follows:
\begin{equation}
    A_{r}=Linear(Concat(A_{prompt},A_{l},A_{s})) ,
\end{equation}
where \(A_{r}\) means to re-model the connection relationship of nodes in the traffic network. This optimized road network can improve the computational efficiency and robustness \cite{zonghan2020connecting} of the model. In order to achieve this better, we will sparsely process the existing fusion matrix as follows:
\begin{equation}
    A_{r}^{i j}=\left\{\begin{array}{ll}
A_{r}^{i j}, &  A_{f}^{i j} \in \operatorname{TopK}\left(A_{r}^{i *}, \tau\right) \\
0, &  \text { otherwise }
\end{array}\right.,
\end{equation}
where \(\tau\) denotes the max number of neighbors.

Finally, we use \(A_{r}\) as an auxiliary information for diffusion-based GCN in the process of capturing dynamic spatial correlations. The process of aggregation of neighbor information between nodes is as follows:
\begin{equation}
    H_{g}^{'}= {\textstyle \sum_{k=0}^{K}A_{r}^{k}H_{g}W} ,
\end{equation}
where \(H_{g}^{'}\in\mathbb{R}^{C\times N \times t^{'}}\) is the output of the Fusion-Graph module and \(k\) denotes the diffusion steps.

Fusion-Graph first considers the potential traffic patterns and the dynamic correlation between nodes caused by spatial heterogeneity, and constructs two dynamic matrices from different perspectives. Then, considering the static geographic information of the real world, it adopts the adjacency matrix as the graph prompt information to jointly construct the fusion matrix, and uses the fusion matrix to model the spatial correlation. Moreover, in order to capture spatial features from global and local dimensions and explore the spatial correlation between nodes in a deeper level, FMPESTF performs the construction of fusion matrix in each ST-Comp module.

\noindent\textbf{Spatial-Temporal Interactive Learning.}
In order to make the model more efficient in capturing spatial-temporal correlations, we conduct spatial-temporal interactive learning in two subsequences \(H_{pre}\in \mathbb{R}^{C\times N \times T/2}\) and \(H_{post}\in \mathbb{R}^{C\times N \times T/2}\) which are divided by \(H\):
\begin{equation}
    H_{pre},H_{post}=Split(H) ,
\end{equation}
After data segmentation is completed, as shown in the Figure \ref{fig_sub}, the specific interaction process is divided into two rounds. The first round of interaction is as follows:
\begin{equation}
    H_{pre}{'}=FGraph(AConv(H_{post}))\odot H_{pre} ,
\end{equation}
\begin{equation}
    H_{post}{'}=FGraph(AConv(H_{pre}))\odot H_{post} ,
\end{equation}
where \(\odot\) denotes Hadamard product, \(FGraph\) and \(AConv\) denote the Fusion-Graph module and the Att-Conv module, and the activation function is omitted. The second round of interactive learning takes the output of the first round of interactive learning as the input, the operation can be defined as follows:
\begin{equation}
    H_{pre}{''}=FGraph(AConv(H_{post}^{'}))+ H_{pre}{'} ,
\end{equation}
\begin{equation}
    H_{post}{''}=FGraph(AConv(H_{pre}^{'}))+ H_{post}{'} .
\end{equation}

Through the above spatial-temporal interactive learning strategy, the model allows sub-sequences to learn each other's spatial-temporal features, so as to achieve positive feedback of spatial-temporal correlation mining. By stacking multi-layer ST-Comp modules, FMPESTF can learn the representation of sequences at different granularity levels. Finally, the output subsequences are rearranged into a complete spatial-temporal representation according to the time sequence, and the output is fused with the original input sequence through residual connection to get the output \(H_{e}\) of the whole encoder.

\subsection{Decoder}
As shown in Figure \ref{fig1}, GLU receives the output \(H_{e}\) from the encoder part of the model as its input. Compared with traditional Multi-Layer Perceptron (MLP) \cite{ilya2021mlp-mixer}, the GLU can control the flow of information more flexibly by introducing a gating mechanism to select and fuse spatial-temporal representations \cite{weizhe2022transformer}, thus enhancing the nonlinear capability of the network.The regression layer uses a fully connected network to further process the output of GLU and get the final forecasting results \(Y\in \mathbb{R}^{N \times T}\). The overall process is as follows:
\begin{equation}
    Y=Linear(GLU(H_{e})) .
\end{equation}

In order to improve computational efficiency and reduce error accumulation, we use a non-autoregressive method to generate forecasting results.

\section{Experiment}

\begin{table}[t]   
    \caption{Statistics of the traffic datasets. }
    \label{tab1}
    \resizebox{0.48\textwidth}{!}{
    \begin{tabular}{c|c c c c}
        \toprule
        \hline
        \textbf{Dataset} & \textbf{Nodes} & \textbf{Time interval} & \textbf{Samples} & \textbf{Time range}\\
        \hline
        \hline
        PEMS03&358&5min&26208&09/01/2018-11/30/2018\\
        PEMS04&307&5min&16992&01/01/2018-02/28/2018\\
        PEMS07&883&5min&28224&05/01/2017-08/31/2017\\
        PEMS08&170&5min&17856&07/01/2016-08/31/2016\\        
        \hline
        NYCBike&250&30min&4368&04/01/2016-06/30/2016\\
        NYCTaxi&266&30min&4368&04/01/2016-06/30/2016\\
        \hline
        \bottomrule
    \end{tabular}   
    }
\end{table}

\begin{table*}[t]
  \begin{center}
    \caption{Performance on highway traffic flow dataset. Bold: Best, Underline: Second best.}
    \vspace{-1em}
    \label{tab2}
    \resizebox{0.85\textwidth}{!}{
    \begin{tabular}{c|c c c| c c c| c c c|c c c }
        \toprule
        \hline
        Dataset & \multicolumn{3}{c}{PEMS03} & \multicolumn{3}{c}{PEMS04}& \multicolumn{3}{c}{PEMS07}& \multicolumn{3}{c}{PEMS08}\\
        \hline
        \diagbox{Method}{Metric}&MAE&RMSE&MAPE&MAE&RMSE&MAPE&MAE&RMSE&MAPE&MAE&RMSE&MAPE\\
        \hline
        \hline
        HA&30.08&46.22&28.64\%&38.51&55.75&28.21\%&45.32&65.74&21.56\%&31.99&46.49&20.28\%\\
        
        VAR&23.65&38.26&24.51\%&24.54&38.61&17.24\%&50.22&75.63&32.22\%&19.19&29.81&13.10\%\\
    
        SVR&21.97&35.29&21.51\%&28.70&44.56&19.20\%&32.49&50.22&14.26\%&23.25&36.16&14.64\%\\
        \hline
        DCRNN&15.53&27.18&15.62\%&19.63&31.26&13.59\%&21.16&34.14&9.02\%&15.22&24.17&10.21\%\\
        
        STGCN&15.65&27.31&15.39\%&19.57&31.38&13.44\%&21.74&35.27&9.24\%&16.08&25.39&10.60\%\\

        ASTGCN&17.34&29.56&17.21\%&22.93&35.22&16.56\%&24.01&37.87&10.73\%&18.25&28.06&11.64\%\\

        GWN&14.80&25.88&\underline{14.92}\%&18.54&30.09&12.71\%&19.84&32.86&8.44\%&14.54&\underline{23.67}&9.41\%\\

        MTGNN&14.88&25.24&15.47\%&18.96&31.05&13.65\%&20.98&34.40&9.31\%&15.12&24.23&9.65\%\\

        AGCRN&15.29&26.95&15.15\%&19.83&32.26&12.97\%&20.57&34.40&8.74\%&15.95&25.22&10.09\%\\

        GMAN&16.52&27.18&17.36\%&18.84&30.756&13.25\%&20.97&34.20&9.05\%&14.57&24.71&9.98\%\\

        STSGCN&17.48&29.21&16.78\%&21.19&33.65&13.90\%&24.26&39.03&10.21\%&17.13&26.80&10.96\%\\

        STFGNN&16.77&28.34&16.30\%&19.83&31.88&13.02\%&22.07&35.80&9.21\%&16.64&26.22&10.60\%\\

        STGGODE&16.50&27.84&16.69\%&20.84&32.82&13.77\%&22.59&37.54&10.14\%&16.81&25.97&10.62\%\\

        ASTGNN&\underline{14.78}&25.00&\textbf{14.79}\%&18.60&30.91&\underline{12.36}\%&20.62&34.00&8.86\%&15.00&24.70&9.50\%\\

        DGCRN&14.80&25.94&15.04\%&18.80&30.65&12.82\%&20.48&33.25&9.06\%&14.60&24.16&9.33\%\\

        STG-NCDE&15.57&27.21&14.68\%&19.30&31.46&12.70\%&21.42&34.51&9.01\%&15.67&24.77&9.94\%\\

        DSTAGNN&15.57&27.21&14.68\%&19.30&31.46&12.70\%&21.42&34.51&9.01\%&15.67&24.77&9.94\%\\

        D\(^{2}\)STGNN&14.88&26.01&15.12\%&\underline{18.34}&\textbf{29.93}&12.81\%&19.68&33.19&8.43\%&14.35&24.18&\underline{9.33}\%\\

        MegaCRN&14.81&26.25&15.16\%&18.70&30.52&12.76\%&19.89&\underline{33.12}&8.47\%&14.68&23.68&9.53\%\\

        STIDGCN&14.98&\underline{24.87}&15.45\%&18.44&30.02&12.62\%&\underline{19.66}&32.87&\underline{8.33}\%&\underline{13.78}&23.95&\underline{9.01}\%\\
        
        \hline
        FMPESTF&\textbf{14.75}&\textbf{24.41}&15.11\%&\textbf{18.17}&\underline{29.96}&\textbf{12.20\%}&\textbf{19.24}&\textbf{32.30}&\textbf{8.01\%}&\textbf{13.53}&\textbf{23.21}&\textbf{8.84\%}\\

        \hline
        \bottomrule
    \end{tabular}
    }
  \end{center}
\end{table*}

In this section, we evaluate the performance of FMPESTF on a series of experiments over six real-world datasets. In order to verify the effectiveness of FMPESTF, we try to answer the following questions:

\textbf{Q1:} How effective of FMPESTF on traffic forecasting task?

\textbf{Q2:} How FMPESTF performs compared to baseline methods on the datasets?

\textbf{Q3:} How do the different variants of FMPESTF perform?

\textbf{Q4:} How do hyperparameters affect the experimenyal results?

\subsection{Experimental Settings}

\textbf{Datasets:} We conducted extensive experiments on four highway traffic flow datasets (PEMS03, PEMS04, PEMS07, PEMS08) \cite{chao2020spatial-temporal} and two traffic demand datasets (NYCBike and NYCTaxi) \cite{junchen2021coupled} including drop-off and pick-up, a total of six real traffic datasets. Table \ref{tab1} provides detailed statics for these datasets. The highway traffic flow dataset was collected by Caltrans's Performance Measurement System (PEMS) \cite{chen2000freeway} and summarized into 5 minutes of observations. Traffic demand data set is derived from urban traffic data. The NYCBike dataset contains daily bike-sharing demand data from New York City bike station residents, while the NYCTaxi dataset contains data from New York City taxi ride records. The two datasets were summarized at 30-minute observation intervals. We use data from the past 12 time horizons to predict data from the future 12 time horizons which can achieve multi-step prediction and divide them into training sets, validation sets, and test sets in a ratio of 6:2:2 to maintain consistency with the previous studies \cite{liu2024spatial-temporal}.

\noindent\textbf{Implementations and Metrics:} We construct all the experiments with PyTorch, which are trained and tested on one NVIDIA 4090 GPU with 24G memory. We train our model using Ranger optimizer \cite{less2021ranger21}, and the initial learning rate \(\alpha\) is set to 0.001. The batch size is set to 64 for the highway traffic flow datasets and 16 for the traffic demand datasets. The training epoch is set to 300, and we use an early stop mechanism during training to ensure the model is over-optimized.

We employ three popular metric methods to evaluate our traffic flow forecasting framework FMPESTF: the Mean Average Error (MAE) Root Mean Square Error (RMSE) and Mean Average Percentage Error (MAPE).

\begin{table*}[t]
  \begin{center}
    \caption{Performance on traffic demand dataset. Bold: Best, Underline: Second best.}
    \vspace{-1em}
    \label{tab3}
    \resizebox{0.85\textwidth}{!}{
    \begin{tabular}{c|c c c| c c c| c c c|c c c }
        \toprule
        \hline
        Dataset & \multicolumn{3}{c}{NYCBike Drop-off} & \multicolumn{3}{c}{NYCBike Pick-up}& \multicolumn{3}{c}{NYCTaxi Drop-off}& \multicolumn{3}{c}{ NYCTaxi Pick-up}\\
        \hline
        \diagbox{Method}{Metric} &MAE&RMSE&MAPE&MAE&RMSE&MAPE&MAE&RMSE&MAPE&MAE&RMSE&MAPE\\
        \hline
        
        DCRNN&1.96&2.94&51.42\%&2.09&3.30&54.22\%&5.19&9.63&37.78\%&5.40&9.71&35.09\%\\
        
        STGCN&2.01&3.07&50.45\%&2.08&3.31&53.12\%&5.38&9.60&39.12\%&5.71&10.22&36.51\%\\

        ASTGCN&2.79&4.20&69.88\%&2.76&4.45&64.23\%&6.98&14.70&45.48\%&7.43&13.84&47.96\%\\

        GWN&1.95&2.98&50.40\%&2.04&3.20&\underline{53.08}\%&5.03&8.78&35.63\%&5.43&9.39&37.79\%\\

        MTGNN&1.94&2.91&50.47\%&2.03&3.19&53.73\%&5.02&8.76&37.62\%&5.39&9.41&37.21\%\\

        AGCRN&2.06&3.19&51.91\%&2.16&3.46&56.35\%&5.45&9.56&40.67\%&5.79&10.11&40.40\%\\

        GMAN&2.09&3.00&54.82\%&2.20&3.35&57.34\%&5.09&8.95&\underline{35.00}\%&5.43&9.47&34.39\%\\

        STSGCN&2.73&4.50&57.89\%&2.36&3.73&58.17\%&5.62&10.21&37.92\%&6.19&11.14&39.67\%\\

        ASTGNN&2.24&3.35&57.89\%&2.36&3.37&58.17\%&5.62&10.21&37.92\%&6.19&11.14&39.67\%\\

        DGCRN&1.96&2.93&51.99\%&2.06&3.21&54.06\%&5.14&9.39&35.09\%&5.44&9.82&35.78\%\\

        STG-NCDE&2.28&3.42&60.96\%&2.15&3.97&55.49\%&5.38&9.74&40.45\%&6.24&11.25&43.20\%\\

        D\(^{2}\)STGNN&\underline{1.92}&2.90&51.94\%&\underline{2.02}&\underline{3.18}&53.60\%&\underline{5.01}&\underline{8.74}&35.81\%&\underline{5.32}&\underline{9.12}&35.51\%\\

        MegaCRN&2.18&3.30&61.42\%&2.31&3.59&67.07\%&5.07&9.11&35.08\%&5.47&9.96&35.13\%\\
        
        STIDGCN&1.94&\underline{2.89}&\underline{49.74}\%&2.21&3.33&\underline{52.01}\%&5.07&8.84&35.68\%&5.33&9.30&\underline{33.98}\%\\
        
        \hline
        FMPESTF&\textbf{1.87}&\textbf{2.76}&\textbf{48.78\%}&\textbf{1.98}&\textbf{3.06}&\textbf{51.50}\%&\textbf{5.01}&\textbf{8.75}&\textbf{34.58\%}&\textbf{5.26}&\textbf{9.26}&\textbf{33.62\%}\\

        \hline
        \bottomrule
    \end{tabular}
    }
  \end{center}
\end{table*}

\subsection{Baseline}
To evaluate our structure prompt enhanced adaptive traffic flow forecasting framework FMPESTF, we compare with state-of-the-art baseline methods, which are listed as follows:

\textbf{HA}, \textbf{VAR} \cite{Zivot2006vector}, \textbf{SVR} \cite{drucker1996support} are methods based on statistics and simple machine learning.
\textbf{DCRNN} \cite{yagyang2018diffusion}, \textbf{GWN} \cite{zonghan2019graph}, \textbf{AGCRN} \cite{lei2020adaptive} are methods based on simple GNN and RNN.
\textbf{ASTGCN} \cite{shengnan2019attention}, \textbf{GMAN} \cite{chuanpan2020gman}, \textbf{ASTGNN} \cite{shengnan2022learning}, \textbf{DSTAGNN} \cite{shiyong2022dstagnn} are attention-based methods.
\textbf{STGCN} \cite{bing018spatio-temporal}, \textbf{MTGNN} \cite{zonghan2020connecting}, \textbf{DGCRN} \cite{fuxian2023dynamic}, \textbf{MegaCRN} \cite{renhe2023spatio-temporal} are methods based on graph convolutions.
\textbf{STSGCN} \cite{chao2020spatial-temporal}, \textbf{STFGNN} \cite{mengzhang2021spatial-temporal} ,\textbf{STGGODE} \cite{zheng2021spatial-temporal} ,\textbf{STG-NCDE} \cite{jeongwhan2022graph}, \textbf{D\(^{2}\)STGNN} \cite{zezhi2022decoupled} are methods based on spatial-temporal graphs and differential equations.
\textbf{STIDGCN} \cite{liu2024spatial-temporal} processes the input data using interactive learning strategies.

\subsection{Comparison and Analysis of Results}

To answer the research questions \textbf{Q1} and \textbf{Q2}, we list the main traffic forecast results on the six datasets in Table \ref{tab2} and Table \ref{tab3} and make the following observations:

\ding{182} Our proposed FMPESTF has excellent performance for the following reasons: \((i)\) FMPESTF constructs a fusion matrix from dynamic dimension and static dimension to simulate the dynamic change of traffic and combine real geospatial information. This method, which combines the dynamic graph structure with the predefined static graph structure, enables FMPESTF to capture the complex spatial correlation within the traffic data. \((ii)\) The time module introduces the attention mechanism, which enables the model to explore the deeper temporal correlation.

\ding{183} Although the traditional methods HA, VAR and SVR processe the temporal information of the data, it does not consider the spatial correlation of the data, so the performance is not ideal. The first proposed STGNNs, such as DCRNN and STGCN, modeled spatially correlated rows and were pioneers in the use of predefined graph structures to enhance spatial correlation capture. The adaptive graph structure methods introduced by GWN, MTGNN and AGCRN improve the model performance by using trainable adaptive embedding to model dynamic spatial correlation. However, these methods cannot adapt to changes in the input data. Data-driven dynamic graph construction methods such as DGCRN, D\(^{2}\)STGNN, and MegaCRN are superior to statistical modeling methods for spatial correlations.

\ding{184} Models based on attention mechanisms, such as GMAN and ASTGNN, also show strong performance. This is due to the fact that the attention mechanism efficiently calculates the correlation between different time steps and nodes, thus better capturing short - and long-time spatial-temporal dependencies. Therefore, our FMPESTF introduces attention mechanism, which improves the performance of the model on the traffic forecasting task.
\begin{figure*}[t]
\centering
\includegraphics[width=0.9\textwidth,trim=0 180 25 0]{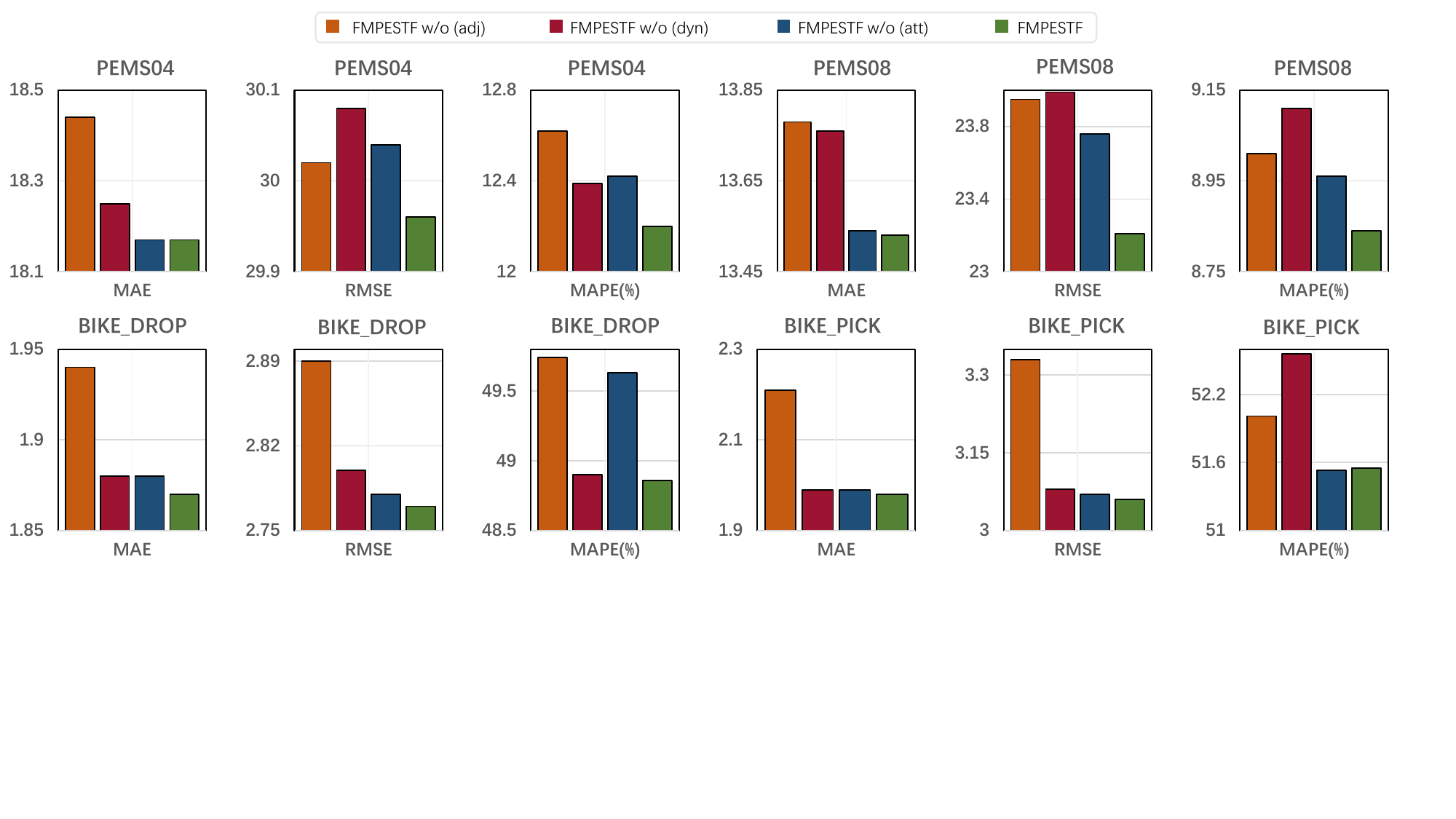}
\caption{Ablation study on three datasets.}
\label{fig4}
\end{figure*}

\subsection{Ablation Studies}

Each module of variants is essential for effective traffic forecasting. In order to answer research question \textbf{Q3}, we conduct an ablation study on the PEMS04, PEMS08, Bike-drop, and Bike-Pick datasets. The variants of our framework variants are listed as follows:
\begin{itemize}
\item \textbf{FMPESTF w/o (adj):} This variant removes the adjacency matrix from the Fusion-Graph module and uses only dynamically generated matrix to diffusion convolution.
\item \textbf{FMPESTF w/o (dyn):} This variant removes the dynamically generated matrix from the Fusion-Graph module and uses only adjacency matrix to diffusion convolution.
\item \textbf{FMPESTF w/o (att):} This variant removes the attention mechanism from the ST-Comp module and uses only 2D convolution.
\end{itemize}
Partial results on four datasets are shown in Figure \ref{fig4}, and we draw the folloewing observation:

\ding{182} All the components had positive effects on FMPESTF. The removal of different components of FMPESTF has different effects on the model performance. Each component in the FMPESTF model has a unique purpose and contributes positively to its overall effectiveness. By combining the strengths of each component, FMPESTF is able to deliver better results and improve its performance in downstream tasks. When evaluating the effects of removing different components from the model, it becomes clear that these effects are inconsistent. Although certain components may have a more significant impact on performance than others, and the removal of different components can lead to varying impacts on the effectiveness. However, the experimental results shown in Figure \ref{fig4} indicate that all components within the FMPESTF contribute positively to its performance.

\ding{183} According to the comparison of FMPESTF w/o (dyn), FMPESTF w/o (adj) and FMPESTF experimental results, Fusion-Graph module plays a crucial role in the model. This module introduces the method of prompt learning to combine real geographic information with dynamic traffic information, thus improving the ability of the model to capture spatial correlation.

\subsection{Hyperparameter Analysis}

\begin{figure}[t]
\centering
\includegraphics[width=0.45\textwidth,trim=0 180 500 20]{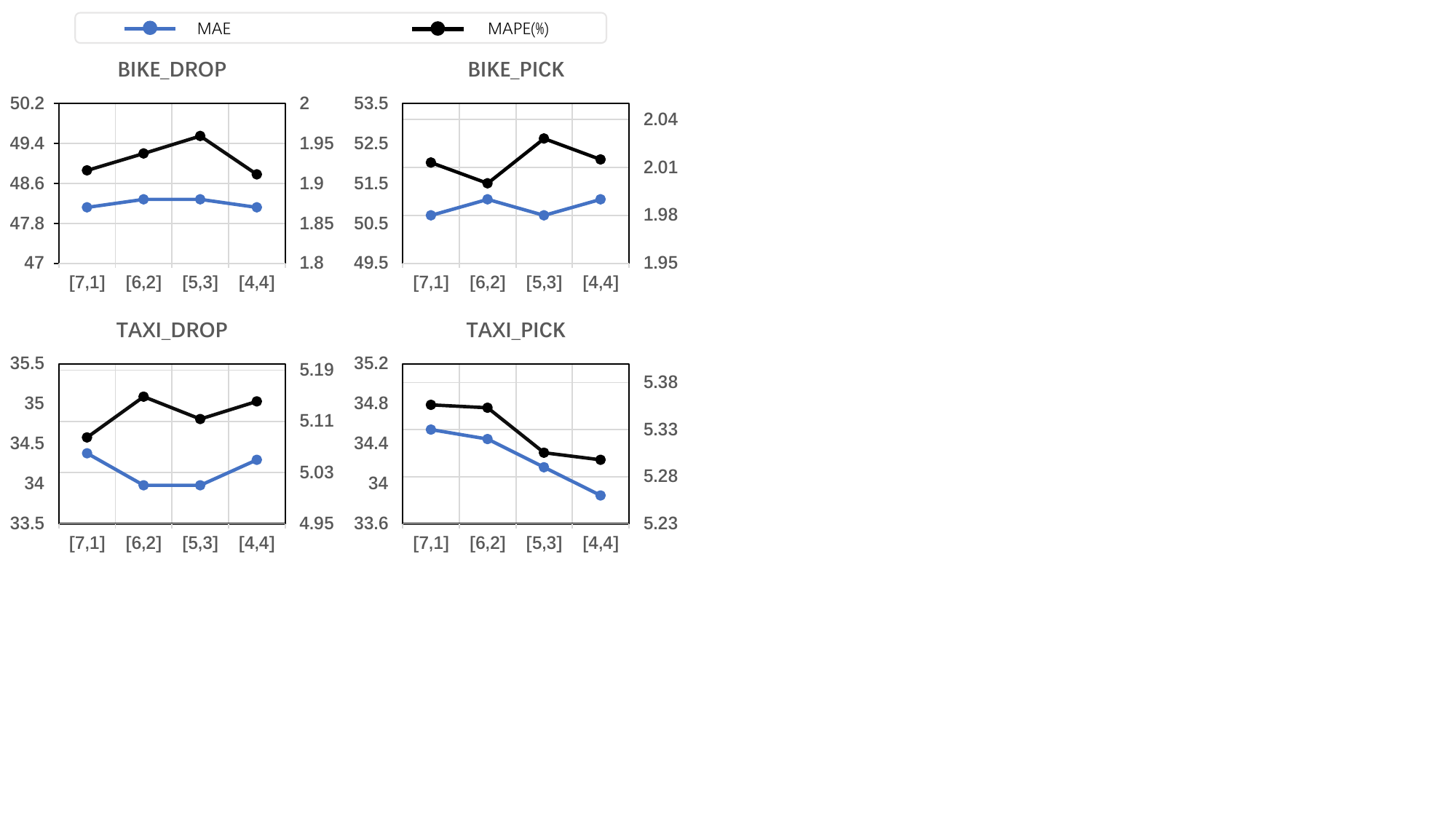}
\caption{Hyperparameter kernel size study on four datasets.  }
\label{fig5}
\end{figure}

\begin{figure}[t]
\centering
\includegraphics[width=0.45\textwidth,trim=0 180 500 20]{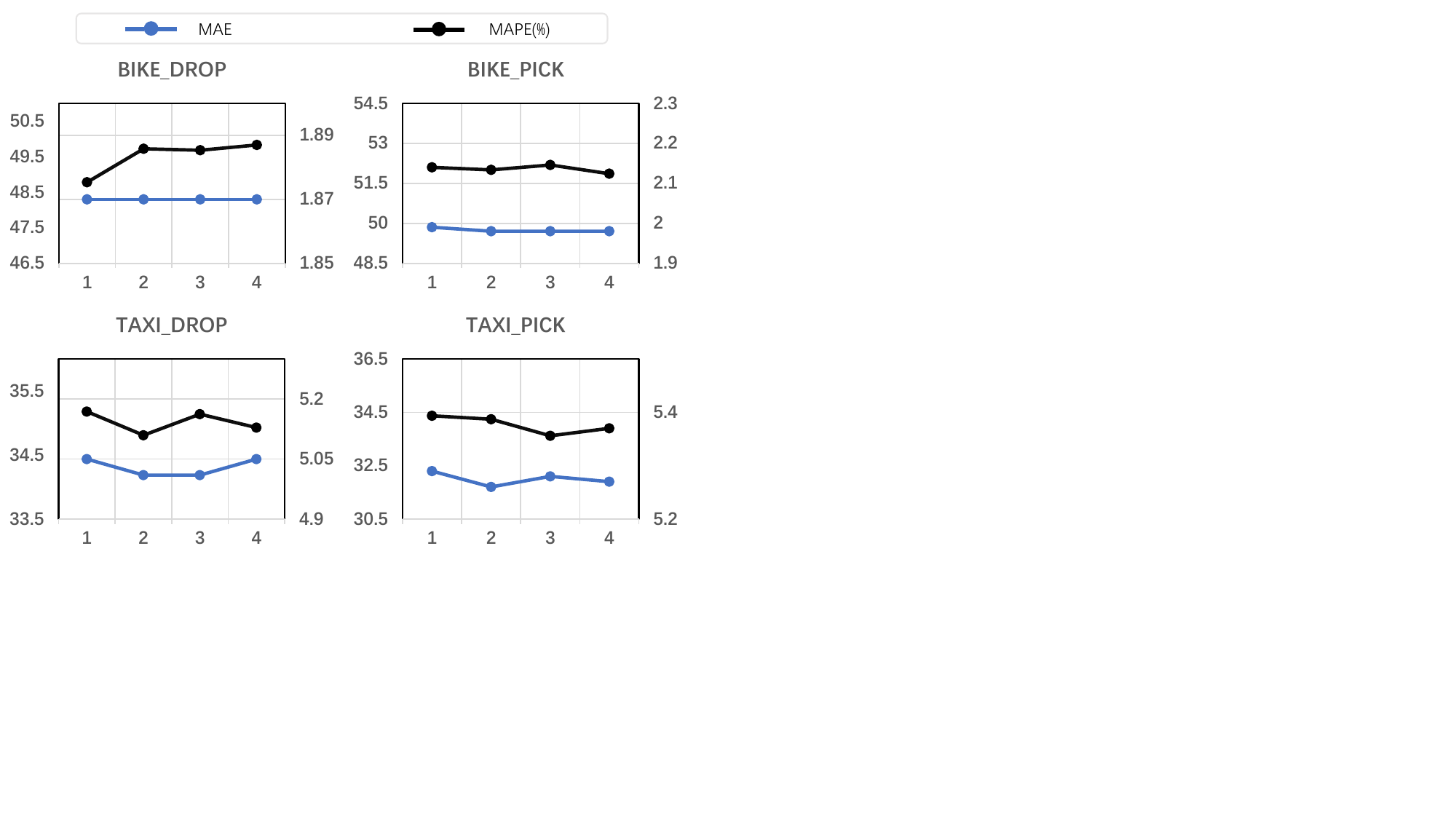}
\caption{Hyperparameter diffusion steps study on four datasets.  }
\label{fig6}
\end{figure}

In order to answer research question \textbf{Q4}, we conducted hyperparameter studies on the Bike-Drop, Bike-Pick, Taxi-Drop and Taxi-Pick datasets. We select the kernel size of the 2D convolution layer in the Att-Conv module and the number of diffusion steps in Fusion-Graph, to study their effects on the model performance. Among these, the kernel size "[7,1]" indicates that the kernel sizes of the two 2D convolution layers in the Att-Conv module are (1,7) and (1,1) respectively. Partial results on four datasets are shown in Figure \ref{fig5} and Figure \ref{fig6}, and the following key insights are derived:

\ding{182} Different kernel size Settings in the convolutional layer enable FMPESTF to sense temporal dependencies at different scales, and the optimal kernel size varies from dataset to dataset.The best choice of kernel size on Bike-Drop and Bike-Pick is [7,1], Taxi-Drop is [5,3], and Taxi-Pick is [4,4].

\ding{183} Increasing the diffusion step size does not obviously improve the performance of FMPESTF, but will increase the calculation cost. The best choice of diffusion steps on Bike-Drop and Bike-Pick is 1, Taxi-Drop and Taxi-Pick is 2. Increasing the diffusion step in diffuse convolutional network allows nodes to sense information from more distant neighbors, thus enhancing the capture of spatial correlations. However, the FMPESTF uses three ST-Comp modules to process the input representation, with each ST-Comp undergoing four rounds of interactive dynamic graph convolution operations. The FMPESTF adds an acceptance domain, similar to increasing the diffusion step, so that nodes can also aggregate information from multi-order neighbors. Therefore, the FMPESTF hyperparameter settings vary across different datasets.

\section{Conclusion}
We propose a fusion matrix prompt enhanced self-attention spatial-temporal interactive traffic forecasting framework named FMPESTF, which consists of three main modules, which combine attention mechanism and fusion matrix with spatial-temporal interactive learning to achieve effective generalized traffic foresting. 
In addition, we introduce an effective Fusion-Graph module based on dynamic Graph construction method, and introduce prompt learning based on adjacency matrix to construct Fusion matrix for spatial correlation modeling. The module makes full use of spatial-temporal information obtained through interactive learning and combines different traffic patterns at each node.
Extensive experiments demonstrate that FMPESTF performs better than baseline on six actual baseline data sets and significantly improves the performance of traffic foresting.

\bibliographystyle{ACM-Reference-Format}


\begin{thebibliography}{46}


\ifx \showCODEN    \undefined \def \showCODEN     #1{\unskip}     \fi
\ifx \showDOI      \undefined \def \showDOI       #1{#1}\fi
\ifx \showISBNx    \undefined \def \showISBNx     #1{\unskip}     \fi
\ifx \showISBNxiii \undefined \def \showISBNxiii  #1{\unskip}     \fi
\ifx \showISSN     \undefined \def \showISSN      #1{\unskip}     \fi
\ifx \showLCCN     \undefined \def \showLCCN      #1{\unskip}     \fi
\ifx \shownote     \undefined \def \shownote      #1{#1}          \fi
\ifx \showarticletitle \undefined \def \showarticletitle #1{#1}   \fi
\ifx \showURL      \undefined \def \showURL       {\relax}        \fi
\providecommand\bibfield[2]{#2}
\providecommand\bibinfo[2]{#2}
\providecommand\natexlab[1]{#1}
\providecommand\showeprint[2][]{arXiv:#2}

\bibitem[Bai et~al\mbox{.}(2020)]%
        {lei2020adaptive}
\bibfield{author}{\bibinfo{person}{Lei Bai}, \bibinfo{person}{Lina Yao}, \bibinfo{person}{Can Li}, \bibinfo{person}{Xianzhi Wang}, {and} \bibinfo{person}{Can Wang}.} \bibinfo{year}{2020}\natexlab{}.
\newblock \showarticletitle{Adaptive Graph Convolutional Recurrent Network for Traffic Forecasting}. In \bibinfo{booktitle}{\emph{Advances in Neural Information Processing Systems 33: Annual Conference on Neural Information Processing Systems 2020, NeurIPS 2020, December 6-12, 2020, virtual}}, \bibfield{editor}{\bibinfo{person}{Hugo Larochelle}, \bibinfo{person}{Marc'Aurelio Ranzato}, \bibinfo{person}{Raia Hadsell}, \bibinfo{person}{Maria{-}Florina Balcan}, {and} \bibinfo{person}{Hsuan{-}Tien Lin}} (Eds.).
\newblock


\bibitem[Chen et~al\mbox{.}(2000)]%
        {chen2000freeway}
\bibfield{author}{\bibinfo{person}{Chao Chen}, \bibinfo{person}{Karl Petty}, {and} \bibinfo{person}{Alex Skabardonis}.} \bibinfo{year}{2000}\natexlab{}.
\newblock \showarticletitle{Freeway performance measurement: Mining loop detector data}.
\newblock \bibinfo{journal}{\emph{Transportation Research Record Journal of the Transportation Research Board}} \bibinfo{number}{1748} (\bibinfo{year}{2000}).
\newblock


\bibitem[Chen et~al\mbox{.}(2024)]%
        {hao2024macro}
\bibfield{author}{\bibinfo{person}{Hao Chen}, \bibinfo{person}{Yuanchen Bei}, \bibinfo{person}{Qijie Shen}, \bibinfo{person}{Yue Xu}, \bibinfo{person}{Sheng Zhou}, \bibinfo{person}{Wenbing Huang}, \bibinfo{person}{Feiran Huang}, \bibinfo{person}{Senzhang Wang}, {and} \bibinfo{person}{Xiao Huang}.} \bibinfo{year}{2024}\natexlab{}.
\newblock \showarticletitle{Macro Graph Neural Networks for Online Billion-Scale Recommender Systems}. In \bibinfo{booktitle}{\emph{Proceedings of the {ACM} on Web Conference 2024, {WWW} 2024, Singapore, May 13-17, 2024}}, \bibfield{editor}{\bibinfo{person}{Tat{-}Seng Chua}, \bibinfo{person}{Chong{-}Wah Ngo}, \bibinfo{person}{Ravi Kumar}, \bibinfo{person}{Hady~W. Lauw}, {and} \bibinfo{person}{Roy~Ka{-}Wei Lee}} (Eds.). \bibinfo{publisher}{{ACM}}, \bibinfo{pages}{3598--3608}.
\newblock


\bibitem[Choi et~al\mbox{.}(2022)]%
        {jeongwhan2022graph}
\bibfield{author}{\bibinfo{person}{Jeongwhan Choi}, \bibinfo{person}{Hwangyong Choi}, \bibinfo{person}{Jeehyun Hwang}, {and} \bibinfo{person}{Noseong Park}.} \bibinfo{year}{2022}\natexlab{}.
\newblock \showarticletitle{Graph Neural Controlled Differential Equations for Traffic Forecasting}. In \bibinfo{booktitle}{\emph{Thirty-Sixth {AAAI} Conference on Artificial Intelligence, {AAAI} 2022, Thirty-Fourth Conference on Innovative Applications of Artificial Intelligence, {IAAI} 2022, The Twelveth Symposium on Educational Advances in Artificial Intelligence, {EAAI} 2022 Virtual Event, February 22 - March 1, 2022}}. \bibinfo{publisher}{{AAAI} Press}, \bibinfo{pages}{6367--6374}.
\newblock


\bibitem[Dauphin et~al\mbox{.}(2017)]%
        {yann2017language}
\bibfield{author}{\bibinfo{person}{Yann~N. Dauphin}, \bibinfo{person}{Angela Fan}, \bibinfo{person}{Michael Auli}, {and} \bibinfo{person}{David Grangier}.} \bibinfo{year}{2017}\natexlab{}.
\newblock \showarticletitle{Language Modeling with Gated Convolutional Networks}. In \bibinfo{booktitle}{\emph{Proceedings of the 34th International Conference on Machine Learning, {ICML} 2017, Sydney, NSW, Australia, 6-11 August 2017}} \emph{(\bibinfo{series}{Proceedings of Machine Learning Research}, Vol.~\bibinfo{volume}{70})}, \bibfield{editor}{\bibinfo{person}{Doina Precup} {and} \bibinfo{person}{Yee~Whye Teh}} (Eds.). \bibinfo{publisher}{{PMLR}}, \bibinfo{pages}{933--941}.
\newblock


\bibitem[Drucker et~al\mbox{.}(1996)]%
        {drucker1996support}
\bibfield{author}{\bibinfo{person}{Harris Drucker}, \bibinfo{person}{Chris J.~C. Burges}, \bibinfo{person}{Linda Kaufman}, \bibinfo{person}{Alex Smola}, {and} \bibinfo{person}{Vladimir Vapnik}.} \bibinfo{year}{1996}\natexlab{}.
\newblock \showarticletitle{Support vector regression machines}. In \bibinfo{booktitle}{\emph{Proceedings of the 9th International Conference on Neural Information Processing Systems}} (Denver, Colorado) \emph{(\bibinfo{series}{NIPS'96})}. \bibinfo{publisher}{MIT Press}, \bibinfo{address}{Cambridge, MA, USA}, \bibinfo{pages}{155–161}.
\newblock


\bibitem[E.Zivot and Wang(2006)]%
        {Zivot2006vector}
\bibfield{author}{\bibinfo{person}{E.Zivot} {and} \bibinfo{person}{J. Wang}.} \bibinfo{year}{2006}\natexlab{}.
\newblock \bibinfo{booktitle}{\emph{Vector Autoregressive Models for Multivariate Time Series}}.
\newblock \bibinfo{publisher}{Springer New York}, \bibinfo{address}{New York, NY}, \bibinfo{pages}{385--429}.
\newblock


\bibitem[Fang et~al\mbox{.}(2023)]%
        {taoran2023universal}
\bibfield{author}{\bibinfo{person}{Taoran Fang}, \bibinfo{person}{Yunchao Zhang}, \bibinfo{person}{Yang Yang}, \bibinfo{person}{Chunping Wang}, {and} \bibinfo{person}{Lei Chen}.} \bibinfo{year}{2023}\natexlab{}.
\newblock \showarticletitle{Universal Prompt Tuning for Graph Neural Networks}. In \bibinfo{booktitle}{\emph{Advances in Neural Information Processing Systems 36: Annual Conference on Neural Information Processing Systems 2023, NeurIPS 2023, New Orleans, LA, USA, December 10 - 16, 2023}}, \bibfield{editor}{\bibinfo{person}{Alice Oh}, \bibinfo{person}{Tristan Naumann}, \bibinfo{person}{Amir Globerson}, \bibinfo{person}{Kate Saenko}, \bibinfo{person}{Moritz Hardt}, {and} \bibinfo{person}{Sergey Levine}} (Eds.).
\newblock


\bibitem[Fang et~al\mbox{.}(2021)]%
        {zheng2021spatial-temporal}
\bibfield{author}{\bibinfo{person}{Zheng Fang}, \bibinfo{person}{Qingqing Long}, \bibinfo{person}{Guojie Song}, {and} \bibinfo{person}{Kunqing Xie}.} \bibinfo{year}{2021}\natexlab{}.
\newblock \showarticletitle{Spatial-Temporal Graph {ODE} Networks for Traffic Flow Forecasting}. In \bibinfo{booktitle}{\emph{{KDD} '21: The 27th {ACM} {SIGKDD} Conference on Knowledge Discovery and Data Mining, Virtual Event, Singapore, August 14-18, 2021}}, \bibfield{editor}{\bibinfo{person}{Feida Zhu}, \bibinfo{person}{Beng~Chin Ooi}, {and} \bibinfo{person}{Chunyan Miao}} (Eds.). \bibinfo{publisher}{{ACM}}, \bibinfo{pages}{364--373}.
\newblock


\bibitem[Guo et~al\mbox{.}(2019)]%
        {shengnan2019attention}
\bibfield{author}{\bibinfo{person}{Shengnan Guo}, \bibinfo{person}{Youfang Lin}, \bibinfo{person}{Ning Feng}, \bibinfo{person}{Chao Song}, {and} \bibinfo{person}{Huaiyu Wan}.} \bibinfo{year}{2019}\natexlab{}.
\newblock \showarticletitle{Attention Based Spatial-Temporal Graph Convolutional Networks for Traffic Flow Forecasting}. In \bibinfo{booktitle}{\emph{The Thirty-Third {AAAI} Conference on Artificial Intelligence, {AAAI} 2019, The Thirty-First Innovative Applications of Artificial Intelligence Conference, {IAAI} 2019, The Ninth {AAAI} Symposium on Educational Advances in Artificial Intelligence, {EAAI} 2019, Honolulu, Hawaii, USA, January 27 - February 1, 2019}}. \bibinfo{publisher}{{AAAI} Press}, \bibinfo{pages}{922--929}.
\newblock


\bibitem[Guo et~al\mbox{.}(2022)]%
        {shengnan2022learning}
\bibfield{author}{\bibinfo{person}{Shengnan Guo}, \bibinfo{person}{Youfang Lin}, \bibinfo{person}{Huaiyu Wan}, \bibinfo{person}{Xiucheng Li}, {and} \bibinfo{person}{Gao Cong}.} \bibinfo{year}{2022}\natexlab{}.
\newblock \showarticletitle{Learning Dynamics and Heterogeneity of Spatial-Temporal Graph Data for Traffic Forecasting}.
\newblock \bibinfo{journal}{\emph{{IEEE} Trans. Knowl. Data Eng.}} \bibinfo{volume}{34}, \bibinfo{number}{11} (\bibinfo{year}{2022}), \bibinfo{pages}{5415--5428}.
\newblock


\bibitem[Hua et~al\mbox{.}(2022)]%
        {weizhe2022transformer}
\bibfield{author}{\bibinfo{person}{Weizhe Hua}, \bibinfo{person}{Zihang Dai}, \bibinfo{person}{Hanxiao Liu}, {and} \bibinfo{person}{Quoc~V. Le}.} \bibinfo{year}{2022}\natexlab{}.
\newblock \showarticletitle{Transformer Quality in Linear Time}. In \bibinfo{booktitle}{\emph{International Conference on Machine Learning, {ICML} 2022, 17-23 July 2022, Baltimore, Maryland, {USA}}} \emph{(\bibinfo{series}{Proceedings of Machine Learning Research}, Vol.~\bibinfo{volume}{162})}, \bibfield{editor}{\bibinfo{person}{Kamalika Chaudhuri}, \bibinfo{person}{Stefanie Jegelka}, \bibinfo{person}{Le~Song}, \bibinfo{person}{Csaba Szepesv{\'{a}}ri}, \bibinfo{person}{Gang Niu}, {and} \bibinfo{person}{Sivan Sabato}} (Eds.). \bibinfo{publisher}{{PMLR}}, \bibinfo{pages}{9099--9117}.
\newblock


\bibitem[Jiang et~al\mbox{.}(2023)]%
        {renhe2023spatio-temporal}
\bibfield{author}{\bibinfo{person}{Renhe Jiang}, \bibinfo{person}{Zhaonan Wang}, \bibinfo{person}{Jiawei Yong}, \bibinfo{person}{Puneet Jeph}, \bibinfo{person}{Quanjun Chen}, \bibinfo{person}{Yasumasa Kobayashi}, \bibinfo{person}{Xuan Song}, \bibinfo{person}{Shintaro Fukushima}, {and} \bibinfo{person}{Toyotaro Suzumura}.} \bibinfo{year}{2023}\natexlab{}.
\newblock \showarticletitle{Spatio-Temporal Meta-Graph Learning for Traffic Forecasting}. In \bibinfo{booktitle}{\emph{Thirty-Seventh {AAAI} Conference on Artificial Intelligence, {AAAI} 2023, Thirty-Fifth Conference on Innovative Applications of Artificial Intelligence, {IAAI} 2023, Thirteenth Symposium on Educational Advances in Artificial Intelligence, {EAAI} 2023, Washington, DC, USA, February 7-14, 2023}}, \bibfield{editor}{\bibinfo{person}{Brian Williams}, \bibinfo{person}{Yiling Chen}, {and} \bibinfo{person}{Jennifer Neville}} (Eds.). \bibinfo{publisher}{{AAAI} Press}, \bibinfo{pages}{8078--8086}.
\newblock


\bibitem[Jin et~al\mbox{.}(2023)]%
        {di2023trafformer}
\bibfield{author}{\bibinfo{person}{Di Jin}, \bibinfo{person}{Jiayi Shi}, \bibinfo{person}{Rui Wang}, \bibinfo{person}{Yawen Li}, \bibinfo{person}{Yuxiao Huang}, {and} \bibinfo{person}{Yu{-}Bin Yang}.} \bibinfo{year}{2023}\natexlab{}.
\newblock \showarticletitle{Trafformer: Unify Time and Space in Traffic Prediction}. In \bibinfo{booktitle}{\emph{Thirty-Seventh {AAAI} Conference on Artificial Intelligence, {AAAI} 2023, Thirty-Fifth Conference on Innovative Applications of Artificial Intelligence, {IAAI} 2023, Thirteenth Symposium on Educational Advances in Artificial Intelligence, {EAAI} 2023, Washington, DC, USA, February 7-14, 2023}}, \bibfield{editor}{\bibinfo{person}{Brian Williams}, \bibinfo{person}{Yiling Chen}, {and} \bibinfo{person}{Jennifer Neville}} (Eds.). \bibinfo{publisher}{{AAAI} Press}, \bibinfo{pages}{8114--8122}.
\newblock


\bibitem[Kong et~al\mbox{.}(2024)]%
        {weiyang2024spatiotemporal}
\bibfield{author}{\bibinfo{person}{Weiyang Kong}, \bibinfo{person}{Ziyu Guo}, {and} \bibinfo{person}{Yubao Liu}.} \bibinfo{year}{2024}\natexlab{}.
\newblock \showarticletitle{Spatio-Temporal Pivotal Graph Neural Networks for Traffic Flow Forecasting}. In \bibinfo{booktitle}{\emph{Thirty-Eighth {AAAI} Conference on Artificial Intelligence, {AAAI} 2024, Thirty-Sixth Conference on Innovative Applications of Artificial Intelligence, {IAAI} 2024, Fourteenth Symposium on Educational Advances in Artificial Intelligence, {EAAI} 2014, February 20-27, 2024, Vancouver, Canada}}, \bibfield{editor}{\bibinfo{person}{Michael~J. Wooldridge}, \bibinfo{person}{Jennifer~G. Dy}, {and} \bibinfo{person}{Sriraam Natarajan}} (Eds.). \bibinfo{publisher}{{AAAI} Press}, \bibinfo{pages}{8627--8635}.
\newblock


\bibitem[Lan et~al\mbox{.}(2022)]%
        {shiyong2022dstagnn}
\bibfield{author}{\bibinfo{person}{Shiyong Lan}, \bibinfo{person}{Yitong Ma}, \bibinfo{person}{Weikang Huang}, \bibinfo{person}{Wenwu Wang}, \bibinfo{person}{Hongyu Yang}, {and} \bibinfo{person}{Pyang Li}.} \bibinfo{year}{2022}\natexlab{}.
\newblock \showarticletitle{{DSTAGNN:} Dynamic Spatial-Temporal Aware Graph Neural Network for Traffic Flow Forecasting}. In \bibinfo{booktitle}{\emph{International Conference on Machine Learning, {ICML} 2022, 17-23 July 2022, Baltimore, Maryland, {USA}}} \emph{(\bibinfo{series}{Proceedings of Machine Learning Research}, Vol.~\bibinfo{volume}{162})}, \bibfield{editor}{\bibinfo{person}{Kamalika Chaudhuri}, \bibinfo{person}{Stefanie Jegelka}, \bibinfo{person}{Le~Song}, \bibinfo{person}{Csaba Szepesv{\'{a}}ri}, \bibinfo{person}{Gang Niu}, {and} \bibinfo{person}{Sivan Sabato}} (Eds.). \bibinfo{publisher}{{PMLR}}, \bibinfo{pages}{11906--11917}.
\newblock


\bibitem[Lei et~al\mbox{.}(2022)]%
        {xiaoliang2022modeling}
\bibfield{author}{\bibinfo{person}{Xiaoliang Lei}, \bibinfo{person}{Hao Mei}, \bibinfo{person}{Bin Shi}, {and} \bibinfo{person}{Hua Wei}.} \bibinfo{year}{2022}\natexlab{}.
\newblock \showarticletitle{Modeling Network-level Traffic Flow Transitions on Sparse Data}. In \bibinfo{booktitle}{\emph{{KDD} '22: The 28th {ACM} {SIGKDD} Conference on Knowledge Discovery and Data Mining, Washington, DC, USA, August 14 - 18, 2022}}, \bibfield{editor}{\bibinfo{person}{Aidong Zhang} {and} \bibinfo{person}{Huzefa Rangwala}} (Eds.). \bibinfo{publisher}{{ACM}}, \bibinfo{pages}{835--845}.
\newblock


\bibitem[Li et~al\mbox{.}(2023)]%
        {fuxian2023dynamic}
\bibfield{author}{\bibinfo{person}{Fuxian Li}, \bibinfo{person}{Jie Feng}, \bibinfo{person}{Huan Yan}, \bibinfo{person}{Guangyin Jin}, \bibinfo{person}{Fan Yang}, \bibinfo{person}{Funing Sun}, \bibinfo{person}{Depeng Jin}, {and} \bibinfo{person}{Yong Li}.} \bibinfo{year}{2023}\natexlab{}.
\newblock \showarticletitle{Dynamic Graph Convolutional Recurrent Network for Traffic Prediction: Benchmark and Solution}.
\newblock \bibinfo{journal}{\emph{{ACM} Trans. Knowl. Discov. Data}} \bibinfo{volume}{17}, \bibinfo{number}{1} (\bibinfo{year}{2023}), \bibinfo{pages}{9:1--9:21}.
\newblock


\bibitem[Li and Zhu(2021)]%
        {mengzhang2021spatial-temporal}
\bibfield{author}{\bibinfo{person}{Mengzhang Li} {and} \bibinfo{person}{Zhanxing Zhu}.} \bibinfo{year}{2021}\natexlab{}.
\newblock \showarticletitle{Spatial-Temporal Fusion Graph Neural Networks for Traffic Flow Forecasting}. In \bibinfo{booktitle}{\emph{Thirty-Fifth {AAAI} Conference on Artificial Intelligence, {AAAI} 2021, Thirty-Third Conference on Innovative Applications of Artificial Intelligence, {IAAI} 2021, The Eleventh Symposium on Educational Advances in Artificial Intelligence, {EAAI} 2021, Virtual Event, February 2-9, 2021}}. \bibinfo{publisher}{{AAAI} Press}, \bibinfo{pages}{4189--4196}.
\newblock


\bibitem[Li et~al\mbox{.}(2018)]%
        {yagyang2018diffusion}
\bibfield{author}{\bibinfo{person}{Yaguang Li}, \bibinfo{person}{Rose Yu}, \bibinfo{person}{Cyrus Shahabi}, {and} \bibinfo{person}{Yan Liu}.} \bibinfo{year}{2018}\natexlab{}.
\newblock \showarticletitle{Diffusion Convolutional Recurrent Neural Network: Data-Driven Traffic Forecasting}. In \bibinfo{booktitle}{\emph{6th International Conference on Learning Representations, {ICLR} 2018, Vancouver, BC, Canada, April 30 - May 3, 2018, Conference Track Proceedings}}. \bibinfo{publisher}{OpenReview.net}.
\newblock


\bibitem[Liu and Zhang(2024)]%
        {liu2024spatial-temporal}
\bibfield{author}{\bibinfo{person}{Aoyu Liu} {and} \bibinfo{person}{Yaying Zhang}.} \bibinfo{year}{2024}\natexlab{}.
\newblock \showarticletitle{Spatial–Temporal Dynamic Graph Convolutional Network With Interactive Learning for Traffic Forecasting}.
\newblock \bibinfo{journal}{\emph{IEEE Transactions on Intelligent Transportation Systems}} \bibinfo{volume}{25}, \bibinfo{number}{7} (\bibinfo{year}{2024}), \bibinfo{pages}{7645--7660}.
\newblock


\bibitem[Liu et~al\mbox{.}(2022)]%
        {minhao2022scinet}
\bibfield{author}{\bibinfo{person}{Minhao Liu}, \bibinfo{person}{Ailing Zeng}, \bibinfo{person}{Muxi Chen}, \bibinfo{person}{Zhijian Xu}, \bibinfo{person}{Qiuxia Lai}, \bibinfo{person}{Lingna Ma}, {and} \bibinfo{person}{Qiang Xu}.} \bibinfo{year}{2022}\natexlab{}.
\newblock \showarticletitle{SCINet: Time Series Modeling and Forecasting with Sample Convolution and Interaction}. In \bibinfo{booktitle}{\emph{Advances in Neural Information Processing Systems 35: Annual Conference on Neural Information Processing Systems 2022, NeurIPS 2022, New Orleans, LA, USA, November 28 - December 9, 2022}}, \bibfield{editor}{\bibinfo{person}{Sanmi Koyejo}, \bibinfo{person}{S.~Mohamed}, \bibinfo{person}{A.~Agarwal}, \bibinfo{person}{Danielle Belgrave}, \bibinfo{person}{K.~Cho}, {and} \bibinfo{person}{A.~Oh}} (Eds.).
\newblock


\bibitem[Liu et~al\mbox{.}(2024)]%
        {zheyuan2024can}
\bibfield{author}{\bibinfo{person}{Zheyuan Liu}, \bibinfo{person}{Xiaoxin He}, \bibinfo{person}{Yijun Tian}, {and} \bibinfo{person}{Nitesh~V. Chawla}.} \bibinfo{year}{2024}\natexlab{}.
\newblock \showarticletitle{Can we Soft Prompt LLMs for Graph Learning Tasks?}. In \bibinfo{booktitle}{\emph{Companion Proceedings of the {ACM} on Web Conference 2024, {WWW} 2024, Singapore, Singapore, May 13-17, 2024}}, \bibfield{editor}{\bibinfo{person}{Tat{-}Seng Chua}, \bibinfo{person}{Chong{-}Wah Ngo}, \bibinfo{person}{Roy~Ka{-}Wei Lee}, \bibinfo{person}{Ravi Kumar}, {and} \bibinfo{person}{Hady~W. Lauw}} (Eds.). \bibinfo{publisher}{{ACM}}, \bibinfo{pages}{481--484}.
\newblock


\bibitem[Mensio et~al\mbox{.}(2018)]%
        {martino2018multi-turn}
\bibfield{author}{\bibinfo{person}{Martino Mensio}, \bibinfo{person}{Giuseppe Rizzo}, {and} \bibinfo{person}{Maurizio Morisio}.} \bibinfo{year}{2018}\natexlab{}.
\newblock \showarticletitle{Multi-turn {QA:} {A} {RNN} Contextual Approach to Intent Classification for Goal-oriented Systems}. In \bibinfo{booktitle}{\emph{Companion of the The Web Conference 2018 on The Web Conference 2018, {WWW} 2018, Lyon , France, April 23-27, 2018}}, \bibfield{editor}{\bibinfo{person}{Pierre{-}Antoine Champin}, \bibinfo{person}{Fabien Gandon}, \bibinfo{person}{Mounia Lalmas}, {and} \bibinfo{person}{Panagiotis~G. Ipeirotis}} (Eds.). \bibinfo{publisher}{{ACM}}, \bibinfo{pages}{1075--1080}.
\newblock


\bibitem[Nebauer(1998)]%
        {claus1998evaluation}
\bibfield{author}{\bibinfo{person}{Claus Nebauer}.} \bibinfo{year}{1998}\natexlab{}.
\newblock \showarticletitle{Evaluation of convolutional neural networks for visual recognition}.
\newblock \bibinfo{journal}{\emph{{IEEE} Trans. Neural Networks}} \bibinfo{volume}{9}, \bibinfo{number}{4} (\bibinfo{year}{1998}), \bibinfo{pages}{685--696}.
\newblock


\bibitem[Rahmani et~al\mbox{.}(2023)]%
        {saeed2023graph}
\bibfield{author}{\bibinfo{person}{Saeed Rahmani}, \bibinfo{person}{Asiye Baghbani}, \bibinfo{person}{Nizar Bouguila}, {and} \bibinfo{person}{Zachary Patterson}.} \bibinfo{year}{2023}\natexlab{}.
\newblock \showarticletitle{Graph Neural Networks for Intelligent Transportation Systems: {A} Survey}.
\newblock \bibinfo{journal}{\emph{{IEEE} Trans. Intell. Transp. Syst.}} \bibinfo{volume}{24}, \bibinfo{number}{8} (\bibinfo{year}{2023}), \bibinfo{pages}{8846--8885}.
\newblock


\bibitem[Scarselli et~al\mbox{.}(2009)]%
        {scarselli2009the}
\bibfield{author}{\bibinfo{person}{Franco Scarselli}, \bibinfo{person}{Marco Gori}, \bibinfo{person}{Ah~Chung Tsoi}, \bibinfo{person}{Markus Hagenbuchner}, {and} \bibinfo{person}{Gabriele Monfardini}.} \bibinfo{year}{2009}\natexlab{}.
\newblock \showarticletitle{The Graph Neural Network Model}.
\newblock \bibinfo{journal}{\emph{IEEE Transactions on Neural Networks}} (\bibinfo{year}{2009}).
\newblock


\bibitem[Shao et~al\mbox{.}(2022a)]%
        {zezhi2022spatial-temporal}
\bibfield{author}{\bibinfo{person}{Zezhi Shao}, \bibinfo{person}{Zhao Zhang}, \bibinfo{person}{Fei Wang}, \bibinfo{person}{Wei Wei}, {and} \bibinfo{person}{Yongjun Xu}.} \bibinfo{year}{2022}\natexlab{a}.
\newblock \showarticletitle{Spatial-Temporal Identity: {A} Simple yet Effective Baseline for Multivariate Time Series Forecasting}. In \bibinfo{booktitle}{\emph{Proceedings of the 31st {ACM} International Conference on Information {\&} Knowledge Management, Atlanta, GA, USA, October 17-21, 2022}}, \bibfield{editor}{\bibinfo{person}{Mohammad~Al Hasan} {and} \bibinfo{person}{Li~Xiong}} (Eds.). \bibinfo{publisher}{{ACM}}, \bibinfo{pages}{4454--4458}.
\newblock


\bibitem[Shao et~al\mbox{.}(2022b)]%
        {zezhi2022decoupled}
\bibfield{author}{\bibinfo{person}{Zezhi Shao}, \bibinfo{person}{Zhao Zhang}, \bibinfo{person}{Wei Wei}, \bibinfo{person}{Fei Wang}, \bibinfo{person}{Yongjun Xu}, \bibinfo{person}{Xin Cao}, {and} \bibinfo{person}{Christian~S. Jensen}.} \bibinfo{year}{2022}\natexlab{b}.
\newblock \showarticletitle{Decoupled Dynamic Spatial-Temporal Graph Neural Network for Traffic Forecasting}.
\newblock \bibinfo{journal}{\emph{Proc. {VLDB} Endow.}} \bibinfo{volume}{15}, \bibinfo{number}{11} (\bibinfo{year}{2022}), \bibinfo{pages}{2733--2746}.
\newblock


\bibitem[Song et~al\mbox{.}(2020)]%
        {chao2020spatial-temporal}
\bibfield{author}{\bibinfo{person}{Chao Song}, \bibinfo{person}{Youfang Lin}, \bibinfo{person}{Shengnan Guo}, {and} \bibinfo{person}{Huaiyu Wan}.} \bibinfo{year}{2020}\natexlab{}.
\newblock \showarticletitle{Spatial-Temporal Synchronous Graph Convolutional Networks: {A} New Framework for Spatial-Temporal Network Data Forecasting}. In \bibinfo{booktitle}{\emph{The Thirty-Fourth {AAAI} Conference on Artificial Intelligence, {AAAI} 2020, The Thirty-Second Innovative Applications of Artificial Intelligence Conference, {IAAI} 2020, The Tenth {AAAI} Symposium on Educational Advances in Artificial Intelligence, {EAAI} 2020, New York, NY, USA, February 7-12, 2020}}. \bibinfo{publisher}{{AAAI} Press}, \bibinfo{pages}{914--921}.
\newblock


\bibitem[Sun et~al\mbox{.}(2022)]%
        {mingchen2022gppt}
\bibfield{author}{\bibinfo{person}{Mingchen Sun}, \bibinfo{person}{Kaixiong Zhou}, \bibinfo{person}{Xin He}, \bibinfo{person}{Ying Wang}, {and} \bibinfo{person}{Xin Wang}.} \bibinfo{year}{2022}\natexlab{}.
\newblock \showarticletitle{{GPPT:} Graph Pre-training and Prompt Tuning to Generalize Graph Neural Networks}. In \bibinfo{booktitle}{\emph{{KDD} '22: The 28th {ACM} {SIGKDD} Conference on Knowledge Discovery and Data Mining, Washington, DC, USA, August 14 - 18, 2022}}, \bibfield{editor}{\bibinfo{person}{Aidong Zhang} {and} \bibinfo{person}{Huzefa Rangwala}} (Eds.). \bibinfo{publisher}{{ACM}}, \bibinfo{pages}{1717--1727}.
\newblock


\bibitem[Tolstikhin et~al\mbox{.}(2021)]%
        {ilya2021mlp-mixer}
\bibfield{author}{\bibinfo{person}{Ilya~O. Tolstikhin}, \bibinfo{person}{Neil Houlsby}, \bibinfo{person}{Alexander Kolesnikov}, \bibinfo{person}{Lucas Beyer}, \bibinfo{person}{Xiaohua Zhai}, \bibinfo{person}{Thomas Unterthiner}, \bibinfo{person}{Jessica Yung}, \bibinfo{person}{Andreas Steiner}, \bibinfo{person}{Daniel Keysers}, \bibinfo{person}{Jakob Uszkoreit}, \bibinfo{person}{Mario Lucic}, {and} \bibinfo{person}{Alexey Dosovitskiy}.} \bibinfo{year}{2021}\natexlab{}.
\newblock \showarticletitle{MLP-Mixer: An all-MLP Architecture for Vision}. In \bibinfo{booktitle}{\emph{Advances in Neural Information Processing Systems 34: Annual Conference on Neural Information Processing Systems 2021, NeurIPS 2021, December 6-14, 2021, virtual}}, \bibfield{editor}{\bibinfo{person}{Marc'Aurelio Ranzato}, \bibinfo{person}{Alina Beygelzimer}, \bibinfo{person}{Yann~N. Dauphin}, \bibinfo{person}{Percy Liang}, {and} \bibinfo{person}{Jennifer~Wortman Vaughan}} (Eds.). \bibinfo{pages}{24261--24272}.
\newblock


\bibitem[Vaswani et~al\mbox{.}(2017)]%
        {ashish2017attention}
\bibfield{author}{\bibinfo{person}{Ashish Vaswani}, \bibinfo{person}{Noam Shazeer}, \bibinfo{person}{Niki Parmar}, \bibinfo{person}{Jakob Uszkoreit}, \bibinfo{person}{Llion Jones}, \bibinfo{person}{Aidan~N. Gomez}, \bibinfo{person}{Lukasz Kaiser}, {and} \bibinfo{person}{Illia Polosukhin}.} \bibinfo{year}{2017}\natexlab{}.
\newblock \showarticletitle{Attention is All you Need}. In \bibinfo{booktitle}{\emph{Advances in Neural Information Processing Systems 30: Annual Conference on Neural Information Processing Systems 2017, December 4-9, 2017, Long Beach, CA, {USA}}}, \bibfield{editor}{\bibinfo{person}{Isabelle Guyon}, \bibinfo{person}{Ulrike von Luxburg}, \bibinfo{person}{Samy Bengio}, \bibinfo{person}{Hanna~M. Wallach}, \bibinfo{person}{Rob Fergus}, \bibinfo{person}{S.~V.~N. Vishwanathan}, {and} \bibinfo{person}{Roman Garnett}} (Eds.). \bibinfo{pages}{5998--6008}.
\newblock


\bibitem[Wang et~al\mbox{.}(2018)]%
        {jingyuan2018mutilevel}
\bibfield{author}{\bibinfo{person}{Jingyuan Wang}, \bibinfo{person}{Ze Wang}, \bibinfo{person}{Jianfeng Li}, {and} \bibinfo{person}{Junjie Wu}.} \bibinfo{year}{2018}\natexlab{}.
\newblock \showarticletitle{Multilevel Wavelet Decomposition Network for Interpretable Time Series Analysis}. In \bibinfo{booktitle}{\emph{Proceedings of the 24th {ACM} {SIGKDD} International Conference on Knowledge Discovery {\&} Data Mining, {KDD} 2018, London, UK, August 19-23, 2018}}, \bibfield{editor}{\bibinfo{person}{Yike Guo} {and} \bibinfo{person}{Faisal Farooq}} (Eds.). \bibinfo{publisher}{{ACM}}, \bibinfo{pages}{2437--2446}.
\newblock


\bibitem[Wright and Demeure(2021)]%
        {less2021ranger21}
\bibfield{author}{\bibinfo{person}{Less Wright} {and} \bibinfo{person}{Nestor Demeure}.} \bibinfo{year}{2021}\natexlab{}.
\newblock \showarticletitle{Ranger21: a synergistic deep learning optimizer}.
\newblock \bibinfo{journal}{\emph{CoRR}}  \bibinfo{volume}{abs/2106.13731} (\bibinfo{year}{2021}).
\newblock


\bibitem[Wu et~al\mbox{.}(2020)]%
        {zonghan2020connecting}
\bibfield{author}{\bibinfo{person}{Zonghan Wu}, \bibinfo{person}{Shirui Pan}, \bibinfo{person}{Guodong Long}, \bibinfo{person}{Jing Jiang}, \bibinfo{person}{Xiaojun Chang}, {and} \bibinfo{person}{Chengqi Zhang}.} \bibinfo{year}{2020}\natexlab{}.
\newblock \showarticletitle{Connecting the Dots: Multivariate Time Series Forecasting with Graph Neural Networks}. In \bibinfo{booktitle}{\emph{{KDD} '20: The 26th {ACM} {SIGKDD} Conference on Knowledge Discovery and Data Mining, Virtual Event, CA, USA, August 23-27, 2020}}, \bibfield{editor}{\bibinfo{person}{Rajesh Gupta}, \bibinfo{person}{Yan Liu}, \bibinfo{person}{Jiliang Tang}, {and} \bibinfo{person}{B.~Aditya Prakash}} (Eds.). \bibinfo{publisher}{{ACM}}, \bibinfo{pages}{753--763}.
\newblock


\bibitem[Wu et~al\mbox{.}(2019)]%
        {zonghan2019graph}
\bibfield{author}{\bibinfo{person}{Zonghan Wu}, \bibinfo{person}{Shirui Pan}, \bibinfo{person}{Guodong Long}, \bibinfo{person}{Jing Jiang}, {and} \bibinfo{person}{Chengqi Zhang}.} \bibinfo{year}{2019}\natexlab{}.
\newblock \showarticletitle{Graph WaveNet for Deep Spatial-Temporal Graph Modeling}. In \bibinfo{booktitle}{\emph{Proceedings of the Twenty-Eighth International Joint Conference on Artificial Intelligence, {IJCAI} 2019, Macao, China, August 10-16, 2019}}, \bibfield{editor}{\bibinfo{person}{Sarit Kraus}} (Ed.). \bibinfo{publisher}{ijcai.org}, \bibinfo{pages}{1907--1913}.
\newblock


\bibitem[Xiao et~al\mbox{.}(2020)]%
        {zhe2020traffic}
\bibfield{author}{\bibinfo{person}{Zhe Xiao}, \bibinfo{person}{Xiuju Fu}, \bibinfo{person}{Liye Zhang}, {and} \bibinfo{person}{Rick Siow~Mong Goh}.} \bibinfo{year}{2020}\natexlab{}.
\newblock \showarticletitle{Traffic Pattern Mining and Forecasting Technologies in Maritime Traffic Service Networks: {A} Comprehensive Survey}.
\newblock \bibinfo{journal}{\emph{{IEEE} Trans. Intell. Transp. Syst.}} \bibinfo{volume}{21}, \bibinfo{number}{5} (\bibinfo{year}{2020}), \bibinfo{pages}{1796--1825}.
\newblock


\bibitem[Yang et~al\mbox{.}(2024)]%
        {yuhao2024graphpro}
\bibfield{author}{\bibinfo{person}{Yuhao Yang}, \bibinfo{person}{Lianghao Xia}, \bibinfo{person}{Da Luo}, \bibinfo{person}{Kangyi Lin}, {and} \bibinfo{person}{Chao Huang}.} \bibinfo{year}{2024}\natexlab{}.
\newblock \showarticletitle{GraphPro: Graph Pre-training and Prompt Learning for Recommendation}. In \bibinfo{booktitle}{\emph{Proceedings of the {ACM} on Web Conference 2024, {WWW} 2024, Singapore, May 13-17, 2024}}, \bibfield{editor}{\bibinfo{person}{Tat{-}Seng Chua}, \bibinfo{person}{Chong{-}Wah Ngo}, \bibinfo{person}{Ravi Kumar}, \bibinfo{person}{Hady~W. Lauw}, {and} \bibinfo{person}{Roy~Ka{-}Wei Lee}} (Eds.). \bibinfo{publisher}{{ACM}}, \bibinfo{pages}{3690--3699}.
\newblock


\bibitem[Ye et~al\mbox{.}(2021)]%
        {junchen2021coupled}
\bibfield{author}{\bibinfo{person}{Junchen Ye}, \bibinfo{person}{Leilei Sun}, \bibinfo{person}{Bowen Du}, \bibinfo{person}{Yanjie Fu}, {and} \bibinfo{person}{Hui Xiong}.} \bibinfo{year}{2021}\natexlab{}.
\newblock \showarticletitle{Coupled Layer-wise Graph Convolution for Transportation Demand Prediction}. In \bibinfo{booktitle}{\emph{Thirty-Fifth {AAAI} Conference on Artificial Intelligence, {AAAI} 2021, Thirty-Third Conference on Innovative Applications of Artificial Intelligence, {IAAI} 2021, The Eleventh Symposium on Educational Advances in Artificial Intelligence, {EAAI} 2021, Virtual Event, February 2-9, 2021}}. \bibinfo{publisher}{{AAAI} Press}, \bibinfo{pages}{4617--4625}.
\newblock


\bibitem[Yu et~al\mbox{.}(2018)]%
        {bing018spatio-temporal}
\bibfield{author}{\bibinfo{person}{Bing Yu}, \bibinfo{person}{Haoteng Yin}, {and} \bibinfo{person}{Zhanxing Zhu}.} \bibinfo{year}{2018}\natexlab{}.
\newblock \showarticletitle{Spatio-Temporal Graph Convolutional Networks: {A} Deep Learning Framework for Traffic Forecasting}. In \bibinfo{booktitle}{\emph{Proceedings of the Twenty-Seventh International Joint Conference on Artificial Intelligence, {IJCAI} 2018, July 13-19, 2018, Stockholm, Sweden}}, \bibfield{editor}{\bibinfo{person}{J{\'{e}}r{\^{o}}me Lang}} (Ed.). \bibinfo{publisher}{ijcai.org}, \bibinfo{pages}{3634--3640}.
\newblock


\bibitem[Zaremba et~al\mbox{.}(2014)]%
        {wojciech2014recurrent}
\bibfield{author}{\bibinfo{person}{Wojciech Zaremba}, \bibinfo{person}{Ilya Sutskever}, {and} \bibinfo{person}{Oriol Vinyals}.} \bibinfo{year}{2014}\natexlab{}.
\newblock \showarticletitle{Recurrent Neural Network Regularization}.
\newblock \bibinfo{journal}{\emph{CoRR}}  \bibinfo{volume}{abs/1409.2329} (\bibinfo{year}{2014}).
\newblock


\bibitem[Zhao et~al\mbox{.}(2023)]%
        {yu2023causal}
\bibfield{author}{\bibinfo{person}{Yu Zhao}, \bibinfo{person}{Pan Deng}, \bibinfo{person}{Junting Liu}, \bibinfo{person}{Xiaofeng Jia}, {and} \bibinfo{person}{Mulan Wang}.} \bibinfo{year}{2023}\natexlab{}.
\newblock \showarticletitle{Causal Conditional Hidden Markov Model for Multimodal Traffic Prediction}. In \bibinfo{booktitle}{\emph{Thirty-Seventh {AAAI} Conference on Artificial Intelligence, {AAAI} 2023, Thirty-Fifth Conference on Innovative Applications of Artificial Intelligence, {IAAI} 2023, Thirteenth Symposium on Educational Advances in Artificial Intelligence, {EAAI} 2023, Washington, DC, USA, February 7-14, 2023}}, \bibfield{editor}{\bibinfo{person}{Brian Williams}, \bibinfo{person}{Yiling Chen}, {and} \bibinfo{person}{Jennifer Neville}} (Eds.). \bibinfo{publisher}{{AAAI} Press}, \bibinfo{pages}{4929--4936}.
\newblock


\bibitem[Zheng et~al\mbox{.}(2020)]%
        {chuanpan2020gman}
\bibfield{author}{\bibinfo{person}{Chuanpan Zheng}, \bibinfo{person}{Xiaoliang Fan}, \bibinfo{person}{Cheng Wang}, {and} \bibinfo{person}{Jianzhong Qi}.} \bibinfo{year}{2020}\natexlab{}.
\newblock \showarticletitle{{GMAN:} {A} Graph Multi-Attention Network for Traffic Prediction}. In \bibinfo{booktitle}{\emph{The Thirty-Fourth {AAAI} Conference on Artificial Intelligence, {AAAI} 2020, The Thirty-Second Innovative Applications of Artificial Intelligence Conference, {IAAI} 2020, The Tenth {AAAI} Symposium on Educational Advances in Artificial Intelligence, {EAAI} 2020, New York, NY, USA, February 7-12, 2020}}. \bibinfo{publisher}{{AAAI} Press}, \bibinfo{pages}{1234--1241}.
\newblock


\bibitem[Zhou et~al\mbox{.}(2020)]%
        {jie2020graph}
\bibfield{author}{\bibinfo{person}{Jie Zhou}, \bibinfo{person}{Ganqu Cui}, \bibinfo{person}{Shengding Hu}, \bibinfo{person}{Zhengyan Zhang}, \bibinfo{person}{Cheng Yang}, \bibinfo{person}{Zhiyuan Liu}, \bibinfo{person}{Lifeng Wang}, \bibinfo{person}{Changcheng Li}, {and} \bibinfo{person}{Maosong Sun}.} \bibinfo{year}{2020}\natexlab{}.
\newblock \showarticletitle{Graph neural networks: {A} review of methods and applications}.
\newblock \bibinfo{journal}{\emph{{AI} Open}}  \bibinfo{volume}{1} (\bibinfo{year}{2020}), \bibinfo{pages}{57--81}.
\newblock


\bibitem[Zhu et~al\mbox{.}(2023)]%
        {yun2023sgl-pt}
\bibfield{author}{\bibinfo{person}{Yun Zhu}, \bibinfo{person}{Jianhao Guo}, {and} \bibinfo{person}{Siliang Tang}.} \bibinfo{year}{2023}\natexlab{}.
\newblock \showarticletitle{{SGL-PT:} {A} Strong Graph Learner with Graph Prompt Tuning}.
\newblock \bibinfo{journal}{\emph{CoRR}}  \bibinfo{volume}{abs/2302.12449} (\bibinfo{year}{2023}).
\newblock


\end{thebibliography}

\appendix

\end{document}